\documentclass[10pt,twocolumn,letterpaper]{article}

\usepackage[pagenumbers]{cvpr} % To force page numbers, e.g. for an arXiv version

\usepackage{graphicx}
\usepackage{amsmath}
\usepackage{amssymb}
\usepackage{booktabs}

\usepackage{multirow}
\usepackage{adjustbox}

\usepackage{bm}

\usepackage{amssymb}% http://ctan.org/pkg/amssymb
\usepackage{pifont}% http://ctan.org/pkg/pifont
\newcommand{\cmark}{\ding{51}}%
\newcommand{\xmark}{\ding{55}}%

\definecolor{cvprblue}{rgb}{0.21,0.49,0.74}
\usepackage[pagebackref,breaklinks,colorlinks,allcolors=cvprblue]{hyperref}

\def\paperID{*****} % *** Enter the Paper ID here
\def\confName{CVPR}
\def\confYear{2025}

\title{Face Reconstruction from Face Embeddings \\ using Adapter to a Face Foundation Model
\vspace{-7pt}
}

\author{\vspace{3pt}Hatef Otroshi Shahreza$^{1,2}$, Anjith George$^{1}$,  
	S\'{e}bastien Marcel$^{1,3}$\\
	$^{1}$Idiap Research Institute, Martigny, Switzerland\\
	$^{2}$\'{E}cole Polytechnique F\'{e}d\'{e}rale de Lausanne (EPFL), Lausanne, Switzerland   \\
	$^{3}$Universit\'{e} de Lausanne (UNIL), Lausanne, Switzerland\\
	{\tt\small \{hatef.otroshi,anjith.george,sebastien.marcel\}@idiap.ch}
 \vspace{-15pt}
}

\begin{document}

% \maketitle

\twocolumn[{%
	\renewcommand\twocolumn[1][]{#1}%
	\maketitle
	\vspace{-10pt}
	\begin{center}
		\centering
		\captionsetup{type=figure}
		
		%\begin{figure*}[h]
		%	\centering
		
		\rotatebox[]{90}{\small  Original \hspace{-60 pt}}\hspace{5pt}\hfil
		\begin{subfigure}[b]{0.139\linewidth}
			\centering
			\includegraphics[page=1,width=.95\linewidth, trim={0.1cm 0cm 0cm 0cm},clip]{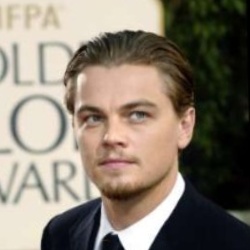}
		\end{subfigure}\hfil
		\begin{subfigure}[b]{0.139\linewidth}
			\centering
			\includegraphics[page=1,width=.95\linewidth, trim={0.4cm 0cm 0.3cm 0.6cm},clip]{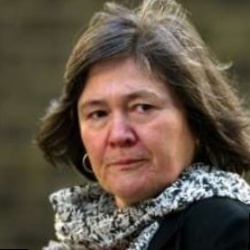}
		\end{subfigure}\hfil
		\begin{subfigure}[b]{0.139\linewidth}
			\centering
			\includegraphics[page=1,width=.95\linewidth, trim={0.3cm 0cm 0.3cm 0.5cm},clip]{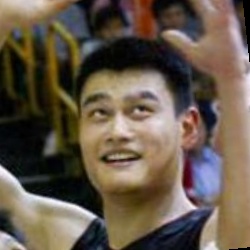}
		\end{subfigure}\hfil
		\begin{subfigure}[b]{0.139\linewidth}
			\centering
			\includegraphics[page=1,width=0.95\linewidth, trim={0.1cm 0cm 0cm 0cm},clip]{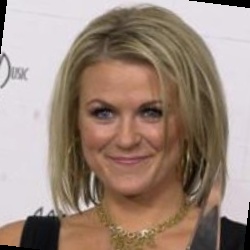}
		\end{subfigure}\hfil
		\begin{subfigure}[b]{0.139\linewidth}
			\centering
			\includegraphics[page=1,width=0.95\linewidth, trim={0.1cm 0cm 0cm 0cm},clip]{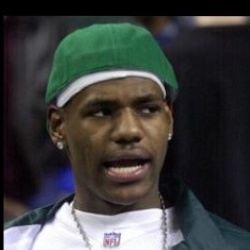}
		\end{subfigure}\hfil
		\begin{subfigure}[b]{0.139\linewidth}
			\centering
			\includegraphics[page=1,width=.95\linewidth, trim={0.1cm 0cm 0cm 0cm},clip]{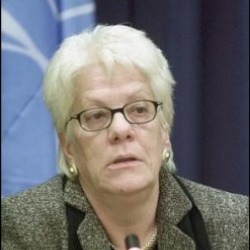}
		\end{subfigure}\hfil
		\begin{subfigure}[b]{0.139\linewidth}
			\centering
			\includegraphics[page=1,width=0.95\linewidth, trim={0.1cm 0cm 0cm 0cm},clip]{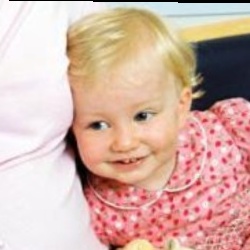}
		\end{subfigure}\hfil \\ \vspace{3pt}
		\rotatebox[]{90}{\small  Reconstruction\hspace{-80 pt}}\hspace{5pt}\hfil
		\begin{subfigure}[b]{0.139\linewidth}
			\centering
			\includegraphics[page=1,width=.95\linewidth, trim={0.1cm 0cm 0cm 0cm},clip]{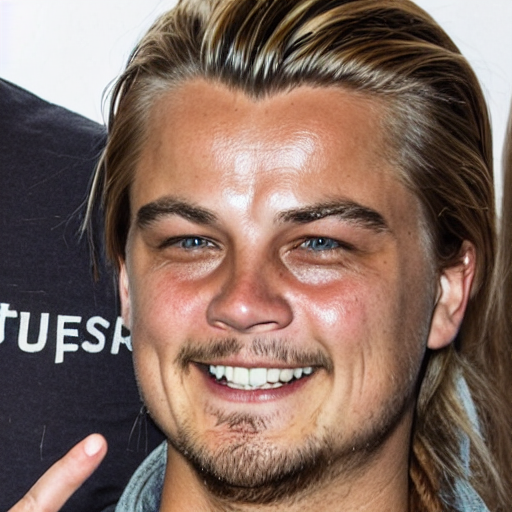}
			\caption*{0.841}
		\end{subfigure}\hfil
		\begin{subfigure}[b]{0.139\linewidth}
			\centering
			\includegraphics[page=1,width=.95\linewidth, trim={0.1cm 0cm 0cm 0cm},clip]{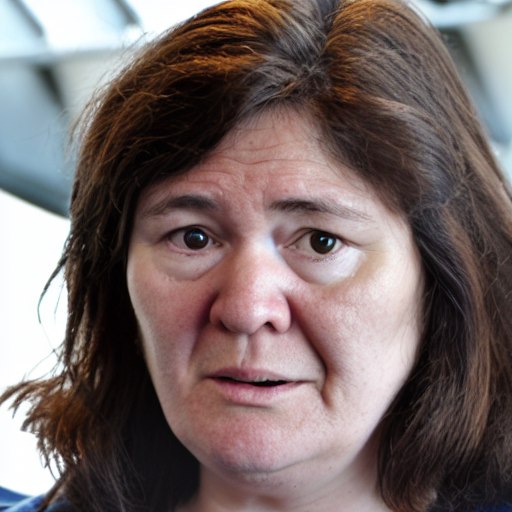}
			\caption*{0.835}
		\end{subfigure}\hfil
		\begin{subfigure}[b]{0.139\linewidth}
			\centering
			\includegraphics[page=1,width=.95\linewidth, trim={0.1cm 0cm 0cm 0cm},clip]{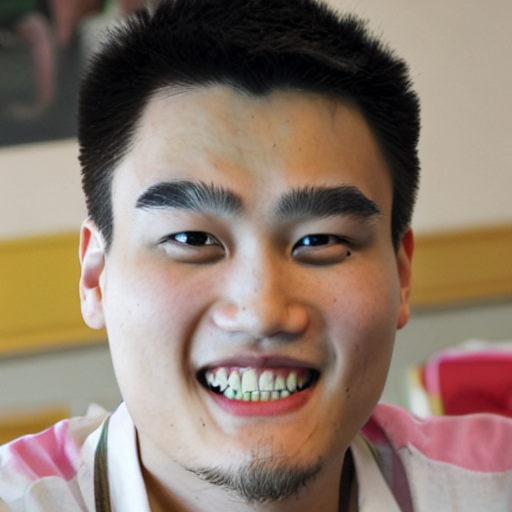}
			\caption*{0.695}
		\end{subfigure}\hfil
		\begin{subfigure}[b]{0.139\linewidth}
			\centering
			\includegraphics[page=1,width=.95\linewidth, trim={0.1cm 0cm 0cm 0cm},clip]{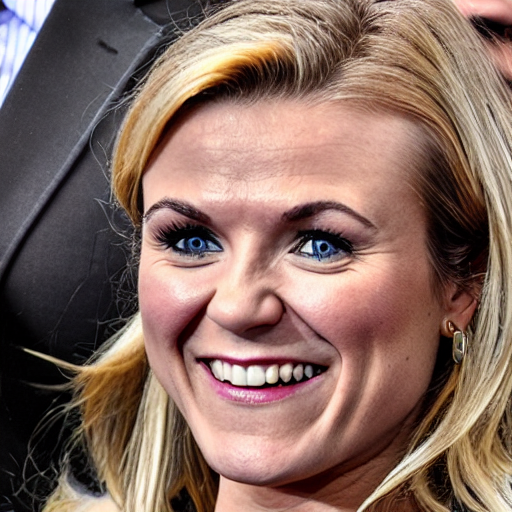}
			\caption*{0.887}
		\end{subfigure}\hfil
		\begin{subfigure}[b]{0.139\linewidth}
			\centering
			\includegraphics[page=1,width=.95\linewidth, trim={0.1cm 0cm 0cm 0cm},clip]{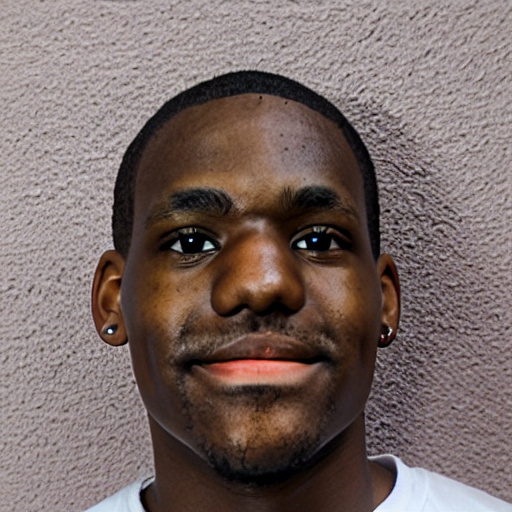}
			\caption*{0.769}
		\end{subfigure}\hfil
		\begin{subfigure}[b]{0.139\linewidth}
			\centering
			\includegraphics[page=1,width=.95\linewidth, trim={0.1cm 0cm 0cm 0cm},clip]{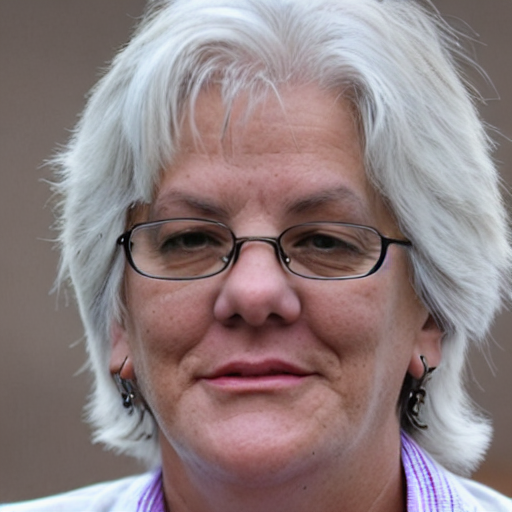}
			\caption*{0.833}
		\end{subfigure}\hfil
		\begin{subfigure}[b]{0.139\linewidth}
			\centering
			\includegraphics[page=1,width=0.95\linewidth, trim={0.1cm 0cm 0cm 0cm},clip]{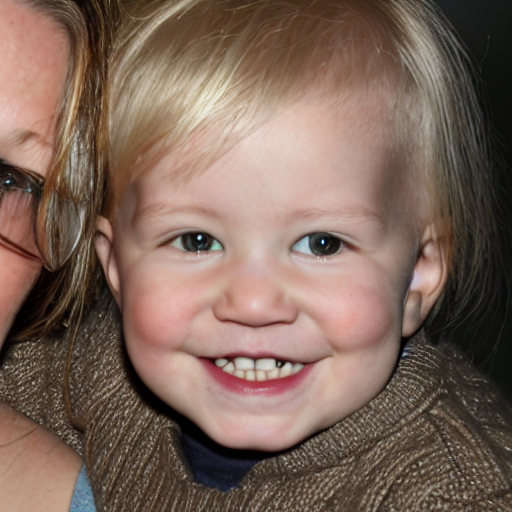}
			\caption*{0.793}
		\end{subfigure}\hfil 
  \vspace{-7pt}
		\caption{Sample face images from the LFW dataset (first row) and their reconstructed versions by our attack (second row) based on ArcFace embeddings. The values are cosine similarity of embeddings of the original and reconstructed face image. %The decision threshold at $\text{FMR}=10^{-3}$ is 0.24 on the LFW dataset. 
		}
		\label{fig:intro:sample_faces}
		%\end{figure*}
		%		\captionof{figure}{Test caption}
	\end{center}%
 \vspace{-3pt}
}]

%%%%%%%%% ABSTRACT
\begin{abstract}
Face recognition systems extract embedding vectors from face images and use these embeddings to verify or identify individuals. Face reconstruction attack (also known as template inversion) refers to reconstructing face images from face embeddings and using the reconstructed face image to enter a face recognition system. In this paper, we propose to use a face foundation model to reconstruct face images from the embeddings of a blackbox face recognition model. The foundation model is trained with 42M images to generate face images from the facial embeddings of a fixed face recognition model. We propose to use an adapter to translate target embeddings into the embedding space of the foundation model. The generated images are evaluated on different face recognition models and different datasets, demonstrating the effectiveness of our method to translate embeddings of different face recognition models. We also evaluate the transferability of reconstructed face images when attacking different face recognition models. Our experimental results show that our reconstructed face images outperform previous reconstruction attacks against face recognition models. \href{https://www.idiap.ch/paper/face_adapter}{Project Page}
\end{abstract}

%%%%%%%%% BODY TEXT
\section{Introduction}\label{sec:intro}
Face recognition systems tend toward ubiquity and are extensively used in various applications that require automatic authentication, ranging from unlocking smartphones to border control. In these systems, a deep neural network is used to extract an embedding vector (also known as a template), which is stored in the database of the system during enrollment. Later during the recognition stage, the extracted embeddings are compared with ones stored in the database of the system. Hence, the extracted embedding vectors play a pivotal role in the automatic recognition of subjects.

With the growing application of face recognition systems, the security of these systems has become a critical topic and attracted considerable attention from the research community \cite{galbally2010vulnerability,biggio2015adversarial,hadid2015biometrics,galbally2014biometric,marcel2023handbook,deb2020advfaces,dong2019efficient}. Among different types of attacks against face recognition systems, face reconstruction attacks (also known as template inversion) threaten the security and privacy of subjects. In a face reconstruction attack, the adversary gains access to face embeddings stored in the database of a face recognition system and tries to reconstruct the face image of enrolled users. The reconstructed face images can provide privacy-sensitive (such as age, gender, ethnicity, etc) information about the user, and in addition, can be used to enter the face recognition system. 
Existing face reconstruction attacks in the literature can be categorized into two groups: learning-based methods and optimization methods. In learning-based methods, such as \cite{cole2017synthesizing,TPAMI2018reconstruction,tpami2023faceti3d}, a network is trained, on a dataset of pairs of embeddings and images, to reconstruct face images from the embeddings. In optimization-based methods, such as \cite{vendrow2021realistic,dong2023reconstruct}, usually a face generator model (such as StyleGAN) is used and through an iterative optimization a latent input is found to generate the reconstructed face images. 
 In each case, the attack requires large computation time for the training (in learning-based methods)  or the inference stage (in optimization-based methods).

Recently, generative models have also absorbed much attention, and there have been significant advancements in the generation capabilities of these models. In a similar vein, different researchers tried to develop large foundation models, which are trained on massive data and can be later used for different applications. 
Recently, the first face foundation model, called Arc2Face~\cite{papantoniou2024arc2face}, was proposed in which the  CLIP \cite{radford2021learning} and Stable Diffusion~\cite{rombach2022high} models were fine-tuned on 42 million face images from WebFace260M dataset~\cite{zhu2021webface260m}. The resulting model is able to generate different face images from embeddings of a fixed face recognition model while preserving identity information. Therefore, the Arc2Face model can be used as a foundation model in many problems which require an identity-conditioned generator. For example, the authors showed that the proposed method achieves state-of-the-art performance in training face recognition using synthetic data. 
However, an important question is that \textit{``given the considerable capabilities of foundation models, 
can face foundation models threaten face recognition systems?"}
% In this paper, we propose a new face reconstruction attack using foundation models. Our method leverages the capability of the foundation model and requires  

In this paper, we focus on face reconstruction attacks and propose a new method to reconstruct face images from different face recognition models using a foundation model. 
We propose an adapter module that can map the face embeddings to the input space of the foundation model. Then, the foundation model can be used to generate face images from the given embeddings. The adapter module enables the use of a foundation model for embeddings of different face recognition models and prevents the need for training a foundation model for a new face recognition model. Therefore, an adversary can easily leverage the capabilities in the foundation model to perform a successful reconstruction attack in a blackbox scenario. 
We perform extensive experiments using several face recognition models on different face recognition datasets. Furthermore, we explore the transferability of reconstructed face images to enter a different face recognition system. The experimental results demonstrate the effectiveness of our reconstructed face images in attacking face recognition systems and superiority compared to previous face reconstruction attacks.  \cref{fig:intro:sample_faces} illustrates sample reconstructed face images by our attack.  

In summary, the contributions of this paper are as follows:
\begin{itemize}
    % \item We propose a new face reconstruction attack against face recognition systems. Our face reconstruction attack is based on a face foundation model that can generate face images from embeddings of a fixed model. However, our attack can reconstruct face images from embeddings of any blackbox model. To our knowledge, this is the first reconstruction attack based on foundation models.
    \item We propose a simple yet effective adapter module to map the face embeddings to the input of the face foundation model. The adapter module enables the use of a foundation model for embeddings of different face recognition models and prevents the need for training a foundation model for a new face recognition model.
    \item We propose a new face reconstruction attack against face recognition systems using a foundation model. While the foundation model can generate face images from embeddings of a fixed model, we use this model to reconstruct face images from embeddings of any blackbox model. 
    To our knowledge, this is the first attack against face recognition systems based on foundation models.
    \item We demonstrate the effectiveness of our reconstructed face images with extensive experiments for different face recognition models on different datasets. We also evaluate the transferability of reconstructed face images for different face recognition models. 
\end{itemize}

In the remainder of the paper, we first review related work in the literature in \cref{sec:related-work}. Next, we propose our face reconstruction method based on a foundation model in \cref{sec:method}. In \cref{sec:experiemnts}, we present our experiments and compare our method with previous attacks in the literature. In \cref{sec:discussion}, we further discuss different aspects of our attack. Finally, the paper is concluded in \cref{sec:conclusion}.

\section{Related Work}\label{sec:related-work}
Previous methods for reconstructing face images from embeddings rely on training a neural network (learning-based) or optimization techniques (optimization-based).
In learning-based methods, a set of face images is used to build a training dataset by extracting their embeddings. Then a network is trained to generate reconstructed face images from embeddings. For example, in \cite{cole2017synthesizing},  a multi-layer perceptron (MLP) and a convolutional neural network (CNN) were trained to generate facial landmark coordinates and face texture, respectively,  from the embeddings. Then,  a differentiable warping was used to combine estimated landmarks and generated textures and reconstruct a low-resolution face image from embedding. 
Their method is proposed for the whitebox face reconstruction attack, where the adversary has full knowledge of the face recognition model (i.e., network structure and its parameters). They further trained MLP and CNN by minimizing the distance between embeddings of reconstructed and original face images. However, for a blackbox attack, they trained  MLP and CNN individually, and used warping in the inference only.
In \cite{TPAMI2018reconstruction}, two new deconvolutional networks, called NbBlock-A and NbBlock-B, are proposed and trained with pixel loss and perceptual loss to generate low-resolution face images from embeddings in blackbox scenarios.
In \cite{duong2020vec2face}, a bijection-learning-based method
is used to train a generative adversarial network (GAN).  
In whitebox attacks, they trained the GAN model by minimizing the distance between embeddings of original and reconstructed images. For blackbox attacks, they used knowledge distillation to train a student network from the face recognition model, and then used the student network in their method. 
In \cite{ahmad2022inverting} a GAN-based face reconstruction network was proposed to generate low-resolution face images in the \textit{blackbox} scenario. However, in contrast to other GAN-based methods,  sample reconstructed face images in their paper do not look realistic and suffer from many artifacts. 
In \cite{shahreza2023blackbox}, a CNN-based face reconstruction network was trained with a proxy face recognition network for blackbox face reconstruction attacks against face recognition systems. Therefore, the reconstructed face images were not realistic and had low resolutions.

Several works in the literature used pretrained face generative models and trained a mapping from face embeddings to different layers of generative models. In \cite{neurips2023faceti}, authors use an adversarial training method to train a mapping from face embeddings to the intermediate layers of StyleGAN3~\cite{StyleGAN3}. While adversarial training leads to an effective mapping, a stable training in an adversarial framework is a difficult task. 
In \cite{dong2021towards}, the authors proposed to generate some face images using StyleGAN2~\cite{StyleGAN2} and extract the embeddings using the face recognition model. Then, they trained an MLP network to map facial embeddings to the input latent codes of StyleGAN2~\cite{StyleGAN2}. In \cite{tbiom2024faceti}, authors used generated images by StyleGAN3~\cite{StyleGAN3} and trained a mapping from face embeddings of synthetic faces to the intermediate layers of StyleGAN3. 
In \cite{tpami2023faceti3d} a semi-supervised learning approach was used to find a mapping to the intermediate latent space of EG3D~\cite{chan2022efficient}. Compared to other methods based on StyleGAN, the method in \cite{tpami2023faceti3d} can generate face images from different points of view thanks to the generative model which is based on neural radiance fields (NeRF). Therefore, the authors proposed camera parameter optimization to find the best pose which increases the attack rate.

\begin{table}[tb]
	\centering
		\renewcommand{\arraystretch}{1.02}
		\setlength{\tabcolsep}{5pt}
		\caption{Comparison with related works.}
		\scalebox{0.78}{
			\begin{tabular}{cccccc}
				\textbf{Ref.}  & \textbf{Resolution}   & \textbf{Method} &  \textbf{Available code}\\ 
				\toprule
				\cite{cole2017synthesizing}  & low & learning-based (MLP+CNN) & \xmark\\ \hline
				\cite{TPAMI2018reconstruction} & low & learning-based (CNN) & \cmark\\ \hline
				\cite{duong2020vec2face}  & low & learning-based (GAN)  & \xmark\\ \hline
                	\cite{ahmad2022inverting}  & low & learning-based (GAN)  & \xmark\\ \hline
                	\cite{shahreza2023blackbox}  & low & learning-based (CNN)  & \cmark\\ \hline
				\cite{neurips2023faceti} & high & learning-based (StyleGAN) & \cmark\\  \hline
				\cite{dong2021towards} & high & learning-based (StyleGAN) & \cmark\\  \hline
				\cite{tbiom2024faceti} & high & learning-based (StyleGAN) & \cmark\\  \hline
				\cite{tpami2023faceti3d} & high & learning-based (EG3D) & \cmark\\  \hline
				\cite{vendrow2021realistic} & high & optimization-based (StyleGAN) & \cmark\\  \hline
				\cite{dong2023reconstruct} & high & optimization-based (StyleGAN) & \cmark\\  \hline
				\textbf{[Ours]} & \textbf{high} & \textbf{learning-based (Arc2Face)} & \cmark\\  
				\bottomrule
			\end{tabular}\label{tab:related-works}
		} 
\end{table}

In contrast to most work in the literature, some papers used optimization techniques to find an input to a pretrained StyleGAN model that can generate the reconstructed face images.
In \cite{vendrow2021realistic}, a grid-search optimization using the simulated annealing \cite{van1987simulated} approach is used on the input of StyleGAN2~\cite{StyleGAN2} that generates the reconstructed face image in blackbox attacks.  A main drawback of optimization-based methods is their high computation for generating face images. This can even cause difficulty in the evaluation of such methods. For example, in \cite{vendrow2021realistic}, authors reported their evaluation on only 20 face images, because of the extensive computation requirement of the method in the inference stage. 
In \cite{dong2023reconstruct}, optimization on the input of StyleGAN2~\cite{StyleGAN2}  was also used,  and the authors used the standard genetic algorithm~\cite{srinivas1994genetic} to solve the optimization.

\cref{tab:related-works} compares our proposed method with the previous works in the literature. All previous methods require large computation, either in the training (learning-based methods) or inference stage (optimization-based methods). However, our method benefits from the capabilities of a foundation model and requires less computation resources for the adversary. We should note that our method also requires training of the adapter module. However, compared to previous learning-based methods which require end-to-end learning, training the adapter module in our method is stable and computationally efficient. 
To our knowledge, our proposed method is the first attack against face recognition systems using face foundation models.

\begin{figure*}[tbh] 
               \centering
               \includegraphics[width=1\textwidth]{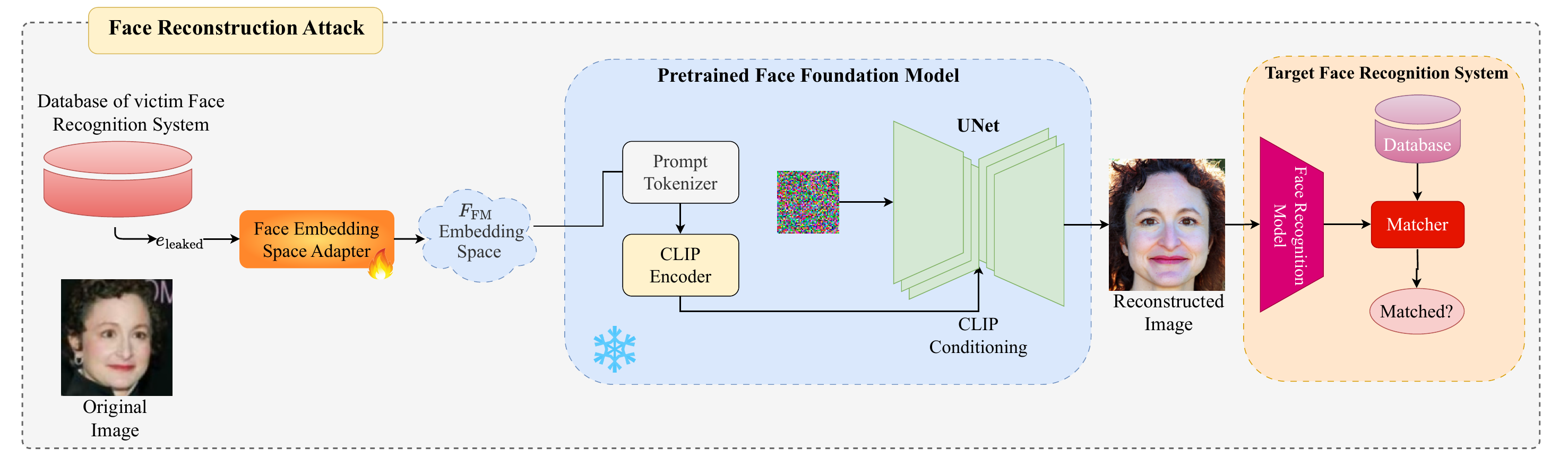}
               \caption{Block diagram for our face reconstruction attack}
               \label{fig:inversionattack} 
\end{figure*}

\section{Face Reconstruction Attack using Foundation Models}\label{sec:method}

\subsection{Threat Model}\label{subsec:method:threat-model}
We consider a face reconstruction attack against a face recognition system based on the following threat model:
\begin{itemize}
	\item \textit{Adversary's goal}: The adversary aims to reconstruct a face image from a leaked face embedding $\bm{e}_\text{leaked}$, and use the reconstructed face image to enter a target face recognition system $F_\text{target}$ system.
	\item \textit{Adversary’s knowledge:} The adversary is assumed to have the following information:
        \begin{itemize}
            \item The adversary knows a face embedding $\bm{e}_\text{leaked}$, which is leaked from database of a victim face recognition system $F_\text{victim}$.
            \item The adversary has a blackbox access (such as SDK, API, etc.) of the feature extractor model $F_\text{victim}$ of the face recognition system from which the embedding $\bm{e}_\text{leaked}$ is leaked.
        \end{itemize} 
	\item \textit{Adversary’s capability:} The adversary can use the reconstructed face image to enter the target face recognition system. For simplicity, we assume that the adversary can inject the reconstructed face image as a query to the target face recognition system. 
	\item \textit{Adversary’s strategy:} The adversary's strategy is to use a foundation model to reconstruct the underlying face image from the leaked embedding $\bm{e}_\text{leaked}$. Then, the adversary can use the reconstructed face images to enter the target face recognition system $F_\text{target}$ (\cref{fig:inversionattack}). 
\end{itemize}

We should note that in our threat model, the victim face recognition system $F_\text{victim}$ from which the embedding $\bm{e}_\text{leaked}$ is leaked can be the same or different than the target face recognition system $F_\text{target}$ system. The latter case (i.e., $F_\text{victim}\neq F_\text{target}$) is a more difficult scenario, where we assume that the same subject is enrolled in another face recognition system with a different feature extractor, and we would like to see if the reconstructed face images from embeddings of $F_\text{victim}$ are \textit{transferable} and can be recognized by a different face recognition model ($F_\text{target}$).

\subsection{Face Reconstruction}\label{subsec:method:face-reconstruction}

As discussed earlier, Arc2Face~\cite{papantoniou2024arc2face} is capable of generating identity-consistent images given an embedding of a specific face recognition model $F_\text{FM}$. Their pipeline includes a modified CLIP~\cite{radford2021learning} model to accept face embeddings from the face recognition model ($F_\text{FM}$) and a Stable Diffusion~\cite{rombach2022high} UNet decoder. This pipeline is end-to-end trained on an upscaled version of WebFace42M~\cite{zhu2021webface260m}, resulting in the generation of high-quality, identity-consistent images. It is evident that this model can be readily used for inverting the embeddings of $F_\text{FM}$, as it is specifically trained for those embeddings. The face recognition model $F_\text{FM}$ used to train Arc2Face has a ResNet100 backbone and is trained with WebFace42M~\cite{zhu2021webface260m}.

\begin{figure}[tb] 
               \centering
               \includegraphics[width=0.99\columnwidth]{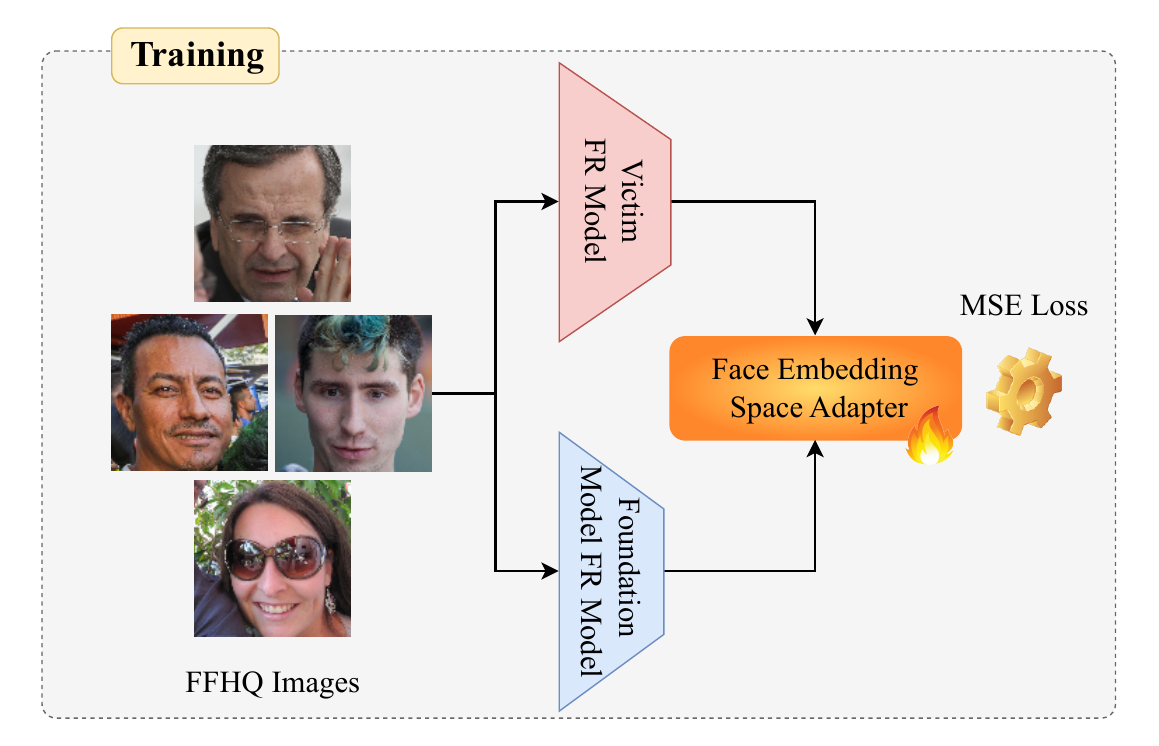}
               \caption{Training process of the adapter module. The parameters of the models are obtained by minimizing MSE loss between embeddings extracted by the (blackbox) victim face recognition model and the face recognition model of the foundation model.}
               \label{fig:training} 
\end{figure}

To attack \textit{any} blackbox face recognition system, one would need to redo the entire process of adapting a large model on a large-scale dataset, which is impractical. In this work, we propose a simple approach to adapt this model for reconstructing images from embeddings of any face recognition system ($F_\text{victim}$).

The main idea is to develop an adapter module that can transform the embeddings from the space of the leaked embedding into the embedding space of $F_\text{FM}$, i.e.,

\begin{equation}
\bm{e}_\text{leaked} \mapsto \bm{e}_\text{FM}
\end{equation}

Essentially, we propose to learn a mapping $\mathcal{M}$ which would map the leaked embeddings to the embedding space of $F_\text{FM}$.

\begin{equation}
\bm{e}_\text{FM} = \mathcal{M}(\bm{e}_\text{leaked})  
\end{equation}

We implement this mapping function, $\mathcal{M}$, as a simple linear layer. The parameters of this adapter layer can be learned using a set of pairs of embeddings extracted from $F_\text{victim}$ and $F_\text{FM}$. %Note that we only need blackbox access to these models to extract the embeddings. %The adapter module can be trained using a set of embeddings from both $F_\text{victim}$ and $F_\text{FM}$. %Specifically, we train the adapter module using embeddings computed from both models using FFHQ images, and the parameters of the adapter are learned through gradient descent with an MSE loss function.
For training the adapter network, we use the Adam \cite{kingma2014adam} optimizer with a learning rate of $10^{-3}$ for 20 epochs, employing Mean Square Error (MSE) as the loss function. It is important to note that we just need blackbox access to the victim model in this process (\cref{fig:training}).

Given the adapter, inverting a leaked embedding from a face recognition system involves first projecting it into the embedding space of $F_\text{FM}$ and then reconstructing it with the Arc2Face foundation model. \cref{fig:inversionattack} depicts our face reconstruction attack.

\section{Experiments}\label{sec:experiemnts}

\subsection{Experimental Setup}
\paragraph{Face Recognition models}
In our experiments, we consider different state-of-the-art face recognition models including 
ArcFace~\cite{deng2018arcface}, ElasticFace~\cite{boutros2021elasticface} with ResNet100 backbones as well as four different face recognition models with state-of-the-art  backbones from FaceX-Zoo~\cite{wang2021facex}, including AttentionNet~\cite{wang2017residual}, HRNet~\cite{wang2020deep}, RepVGG~\cite{ding2021repvgg}
and Swin~\cite{liu2021swin}. \cref{{tab:exp:FR-performance}} reports the recognition performance of these models. 
All of these models are trained on the MS-Celeb1M dataset~\cite{guo2016ms}. Note that the face recognition model $F_\text{FM}$ used in the Arc2Face model is also trained with ArcFace loss function, but on the WebFace42M~\cite{zhu2021webface260m} dataset. In our experiments, we denote the face recognition model used in the foundation model (i.e., trained with WebFace42M) with $F_\text{FM}$, and refer to the one trained with  MS-Celeb1M dataset~\cite{guo2016ms} as ArcFace for a victim or target face recognition model. Although both the models are trained with the ArcFace loss function, the latent spaces of these models are not aligned due to the different initialization and training datasets.

\paragraph{Datasets}
To train the adapter module, we use the Flickr-Faces-HQ (FFHQ) dataset~\cite{StyleGAN}, which consists of 70,000 high-quality images collected from the internet (without identity labels). All 70,000 images from the FFHQ dataset are cropped and aligned using MTCNN \cite{zhang2016joint}. We use 60,000 images for training and reserve the remaining 10,000 as the test set.

To evaluate the vulnerability of face recognition models, we use three benchmarking datasets, including MOBIO~\cite{MOBIO}, Labeled Faces in the Wild~(LFW)~\cite{LFW} and AgeDB~\cite{moschoglou2017agedb} datasets. The MOBIO dataset includes face images of 150 people captured using mobile devices in 12 sessions for each person. The LFW dataset consists of 13,233 face images of 5,749 people collected from the internet, where 1,680 people have two or more images. The  AgeDB~\cite{moschoglou2017agedb} dataset includes 16,488 images (in different ages for each subject) of 568 famous people.

\paragraph{Evaluation Protocol}
For evaluation, we build a face recognition system using one of the mentioned feature extractors as $F_\text{victim}$ and train an adapter module for the corresponding feature extractor. We use the benchmarking datasets for the verification task (i.e., 1:1 matching) and consider the enrolled embedding as leaked embedding which is used by the adversary for reconstruction attack. After reconstruction, we inject the reconstructed face image as a query to the feature extractor of the target system and evaluate the adversary's success attack rate (SAR) to enter the target system. We should note that the target face recognition system may have the same or different (i.e., transferability evaluation) feature extractor model that the system from which the embeddings are leaked. 

\paragraph{Implementation  and Source Code}
Our experiments, including training of adapter and comparison with previous methods, are conducted on a system equipped with a single NVIDIA RTX 3090 GPU. For the face foundation model, we use a pretrained model of Arc2Face~\cite{papantoniou2024arc2face} in our experiments and generate $512\times512$ resolution face images. 
The source code of our experiments is publicly available\footnote{Project page: \href{https://www.idiap.ch/paper/face_adapter}{https://www.idiap.ch/paper/face\_adapter}}.

\begin{table}[t]
	\centering
	\renewcommand{\arraystretch}{1.02}
	\setlength{\tabcolsep}{7pt}
	\caption{Recognition performance of face recognition models in terms of true match rate (TMR) at false match rate (FMR)  of $10^{-3}$ evaluated on the MOBIO, LFW, and AgeDB datasets. 
	}
	\scalebox{0.85}{
		\begin{tabular}{lcccccccccccccc}
			\multirow{1}{*}{\textbf{\scalebox{1}{Face Recognition}}}   &  \multicolumn{1}{c}{\textbf{MOBIO}} &  \multicolumn{1}{c}{\textbf{LFW}} &  \multicolumn{1}{c}{ \textbf{AgeDB}}   \\ 
			% \cline{2-3}
			% \cline{5-6}
			% \cline{8-9} 
			% &   \textbf{FMR=$\mathbf{10^{-2}}$}  & \textbf{FMR=$\mathbf{10^{-3}}$}   &  &   \textbf{FMR=$\mathbf{10^{-2}}$}  & \textbf{ FMR=$\mathbf{10^{-3}}$} &  &    \textbf{FMR=$\mathbf{10^{-2}}$}  &  \textbf{ FMR=$\mathbf{10^{-3}}$}   \\
			\toprule
			\textbf{ArcFace} & 99.98 & 96.40 & 98.07   \\ \hline
			\textbf{ElasticFace} & 100.00 & 94.70 &  97.57  \\ \hline
			\textbf{AttentionNet} &  97.73 & 72.77 & 96.90  \\ \hline
			\textbf{HRNet}  & 98.23 & 78.43  & 96.23  \\ \hline
			\textbf{RepVGG} &  95.80 & 58.07 & 93.93     \\ \hline
			\textbf{Swin}  & 98.98 & 87.83 & 97.10   \\ 
			\bottomrule
		\end{tabular}\label{tab:exp:FR-performance}
	}
\end{table}

\subsection{Analyze}
\begin{figure}[t]
	\centering
	\rotatebox[]{90}{\small  Original \hspace{-45 pt}}\hspace{5pt}\hfil
	\begin{subfigure}[b]{0.22\linewidth}
		\centering
		\includegraphics[page=1,width=.95\linewidth, trim={0.1cm 0cm 0cm 0cm},clip]{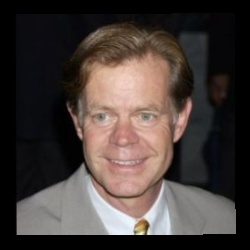}
	\end{subfigure}\hfil
	\begin{subfigure}[b]{0.22\linewidth}
		\centering
		\includegraphics[page=1,width=.95\linewidth, trim={0.1cm 0cm 0cm 0cm},clip]{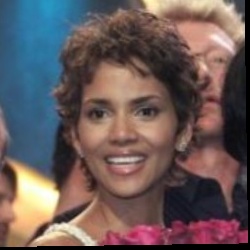}
	\end{subfigure}\hfil
	\begin{subfigure}[b]{0.22\linewidth}
		\centering
		\includegraphics[page=1,width=.95\linewidth, trim={0.1cm 0cm 0cm 0cm},clip]{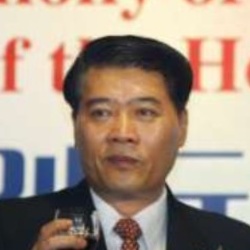}
	\end{subfigure}\hfil
	\begin{subfigure}[b]{0.22\linewidth}
		\centering
		\includegraphics[page=1,width=.95\linewidth, trim={0.1cm 0cm 0cm 0cm},clip]{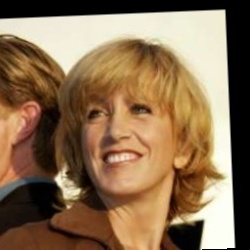}
	\end{subfigure}\hfil
	\\ \vspace{3pt}
	\rotatebox[]{90}{\small  ArcFace \hspace{-70 pt}}\hspace{5pt}\hfil
	\begin{subfigure}[b]{0.22\linewidth}
		\centering
		\includegraphics[page=1,width=0.95\linewidth, trim={0.1cm 0cm 0cm 0cm},clip]{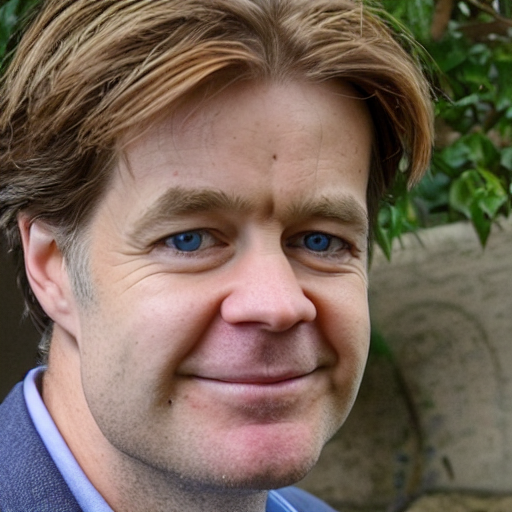}
		\caption*{0.749}
	\end{subfigure}\hfil
	\begin{subfigure}[b]{0.22\linewidth}
		\centering
		\includegraphics[page=1,width=0.95\linewidth, trim={0.1cm 0cm 0cm 0cm},clip]{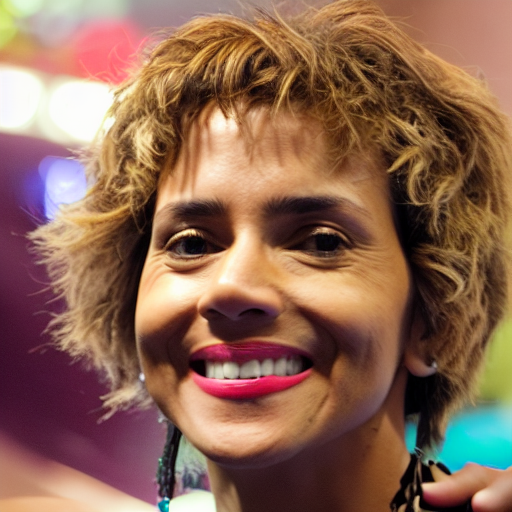}
		\caption*{0.729}
	\end{subfigure}\hfil
	\begin{subfigure}[b]{0.22\linewidth}
		\centering
		\includegraphics[page=1,width=0.95\linewidth, trim={0.1cm 0cm 0cm 0cm},clip]{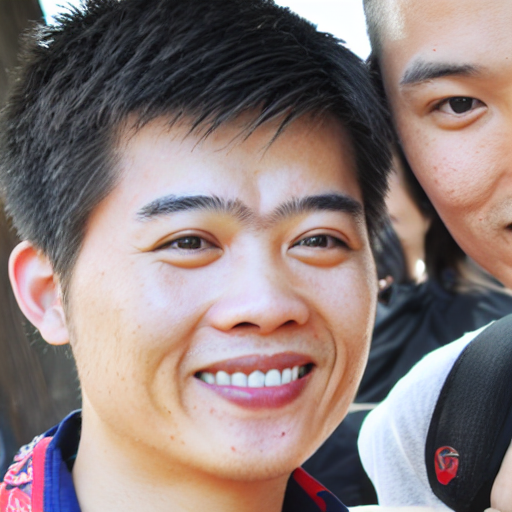}
		\caption*{0.740}
	\end{subfigure}\hfil
	\begin{subfigure}[b]{0.22\linewidth}
		\centering
		\includegraphics[page=1,width=0.95\linewidth, trim={0.1cm 0cm 0cm 0cm},clip]{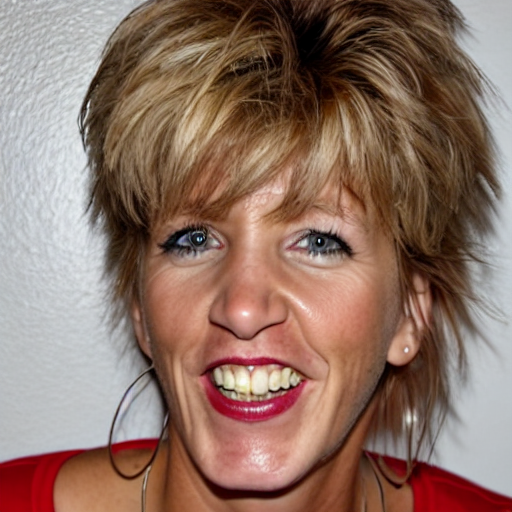}
		\caption*{0.813}
	\end{subfigure}\hfil
	\\ \vspace{3pt}
	\rotatebox[]{90}{\small  ElasticFace \hspace{-70 pt}}\hspace{5pt}\hfil
	\begin{subfigure}[b]{0.22\linewidth}
		\centering
		\includegraphics[page=1,width=0.95\linewidth, trim={0.1cm 0cm 0cm 0cm},clip]{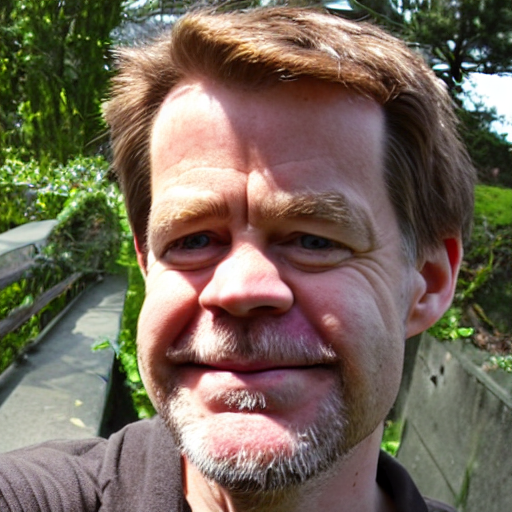}
		\caption*{0.792}
	\end{subfigure}\hfil
	\begin{subfigure}[b]{0.22\linewidth}
		\centering
		\includegraphics[page=1,width=0.95\linewidth, trim={0.1cm 0cm 0cm 0cm},clip]{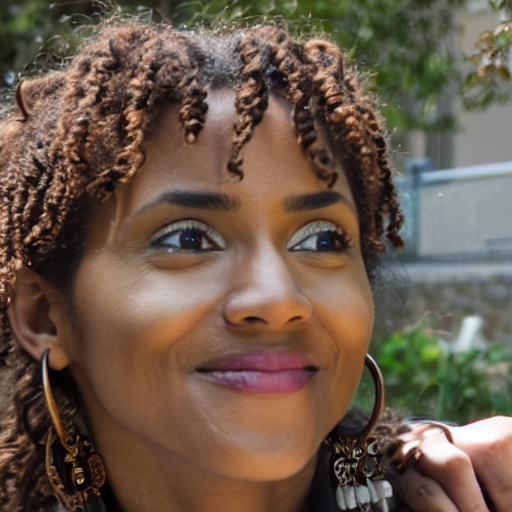}
		\caption*{0.746}
	\end{subfigure}\hfil
	\begin{subfigure}[b]{0.22\linewidth}
		\centering
		\includegraphics[page=1,width=0.95\linewidth, trim={0.1cm 0cm 0cm 0cm},clip]{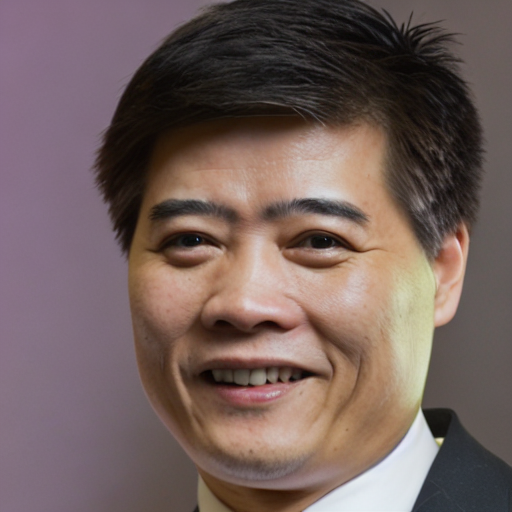}
		\caption*{0.810}
	\end{subfigure}\hfil
	\begin{subfigure}[b]{0.22\linewidth}
		\centering
		\includegraphics[page=1,width=0.95\linewidth, trim={0.1cm 0cm 0cm 0cm},clip]{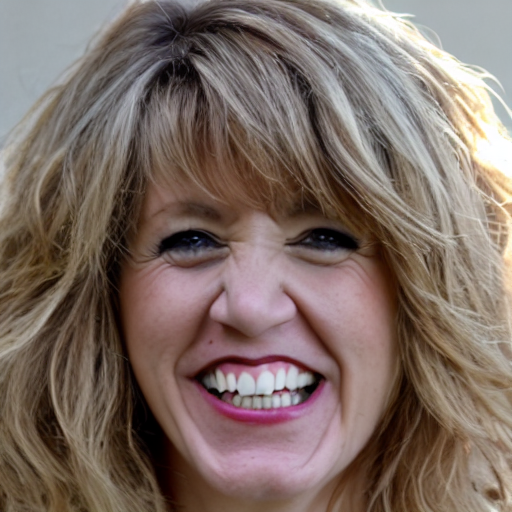}
		\caption*{0.773}
	\end{subfigure}\hfil
	\\ \vspace{3pt}
	\rotatebox[]{90}{\small  AttentionNet \hspace{-70 pt}}\hspace{5pt}\hfil
	\begin{subfigure}[b]{0.22\linewidth}
		\centering
		\includegraphics[page=1,width=0.95\linewidth, trim={0.1cm 0cm 0cm 0cm},clip]{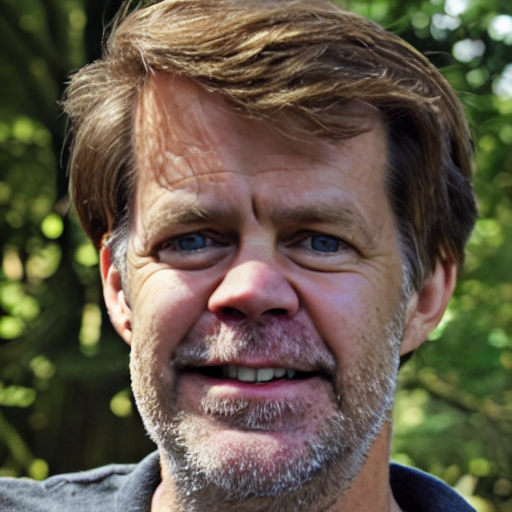}
		\caption*{0.813}
	\end{subfigure}\hfil
	\begin{subfigure}[b]{0.22\linewidth}
		\centering
		\includegraphics[page=1,width=0.95\linewidth, trim={0.1cm 0cm 0cm 0cm},clip]{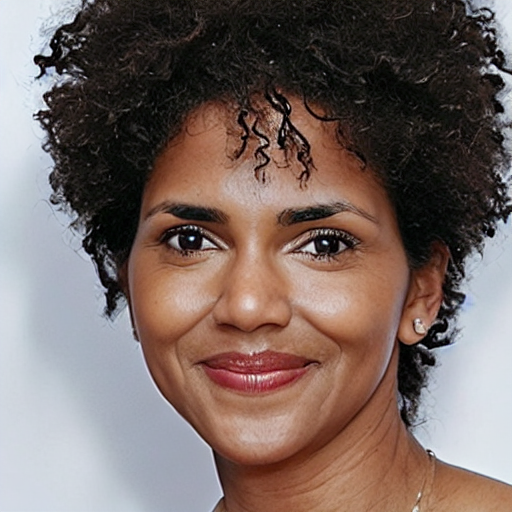}
		\caption*{0.856}
	\end{subfigure}\hfil
	\begin{subfigure}[b]{0.22\linewidth}
		\centering
		\includegraphics[page=1,width=0.95\linewidth, trim={0.1cm 0cm 0cm 0cm},clip]{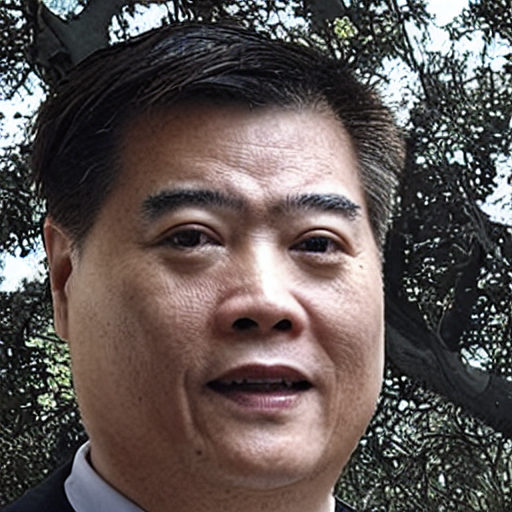}
		\caption*{0.838}
	\end{subfigure}\hfil
	\begin{subfigure}[b]{0.22\linewidth}
		\centering
		\includegraphics[page=1,width=0.95\linewidth, trim={0.1cm 0cm 0cm 0cm},clip]{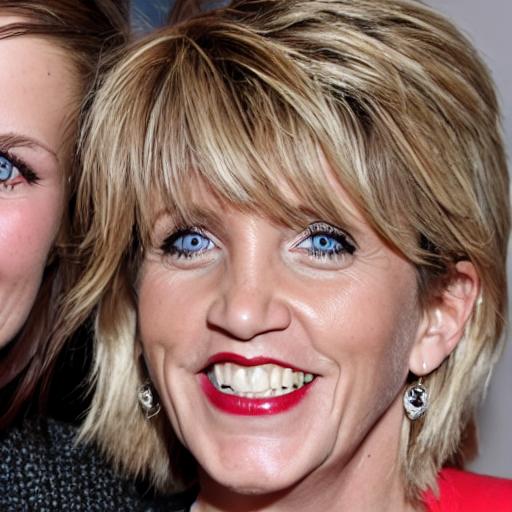}
		\caption*{0.808}
	\end{subfigure}\hfil
	\\ \vspace{3pt}
	\rotatebox[]{90}{\small  HRNet \hspace{-70 pt}}\hspace{5pt}\hfil
	\begin{subfigure}[b]{0.22\linewidth}
		\centering
		\includegraphics[page=1,width=0.95\linewidth, trim={0.1cm 0cm 0cm 0cm},clip]{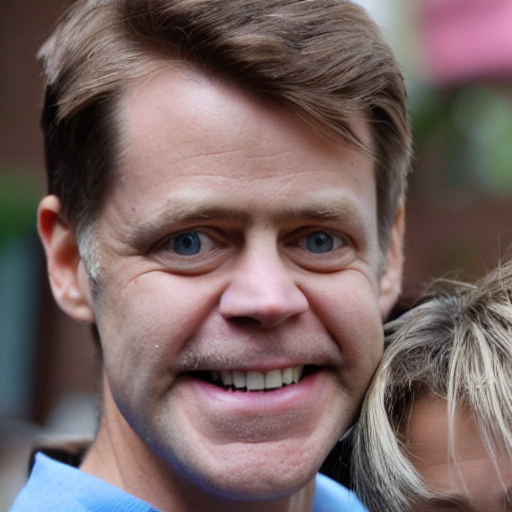}
		\caption*{0.795}
	\end{subfigure}\hfil
	\begin{subfigure}[b]{0.22\linewidth}
		\centering
		\includegraphics[page=1,width=0.95\linewidth, trim={0.1cm 0cm 0cm 0cm},clip]{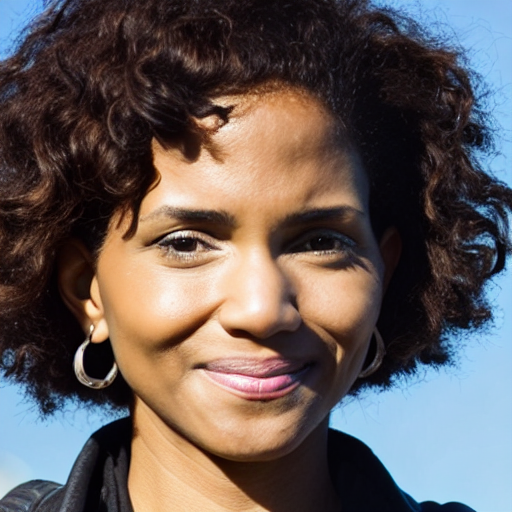}
		\caption*{0.775}
	\end{subfigure}\hfil
	\begin{subfigure}[b]{0.22\linewidth}
		\centering
		\includegraphics[page=1,width=0.95\linewidth, trim={0.1cm 0cm 0cm 0cm},clip]{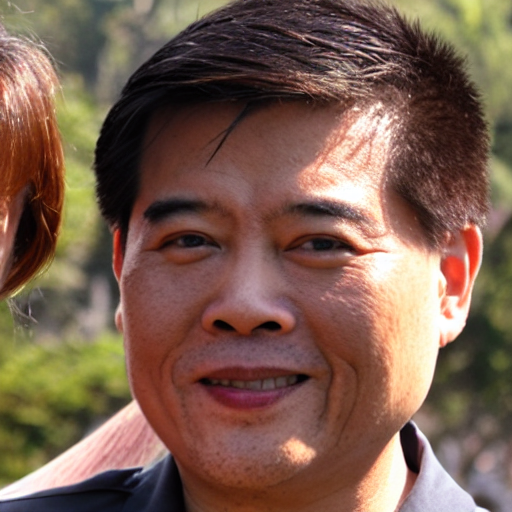}
		\caption*{0.780}
	\end{subfigure}\hfil
	\begin{subfigure}[b]{0.22\linewidth}
		\centering
		\includegraphics[page=1,width=0.95\linewidth, trim={0.1cm 0cm 0cm 0cm},clip]{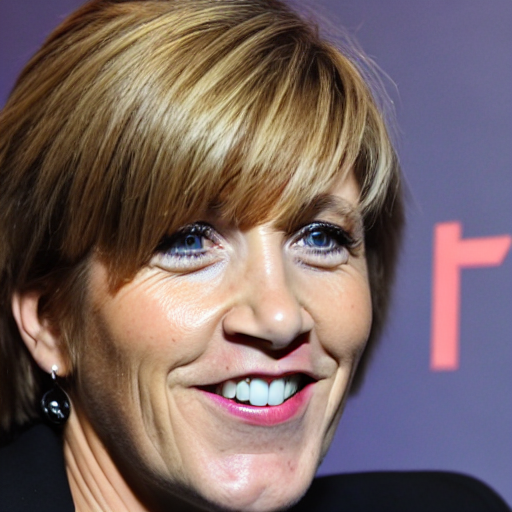}
		\caption*{0.806}
	\end{subfigure}\hfil
	\\ \vspace{3pt}
	\rotatebox[]{90}{\small  RepVGG \hspace{-70 pt}}\hspace{5pt}\hfil
	\begin{subfigure}[b]{0.22\linewidth}
		\centering
		\includegraphics[page=1,width=0.95\linewidth, trim={0.1cm 0cm 0cm 0cm},clip]{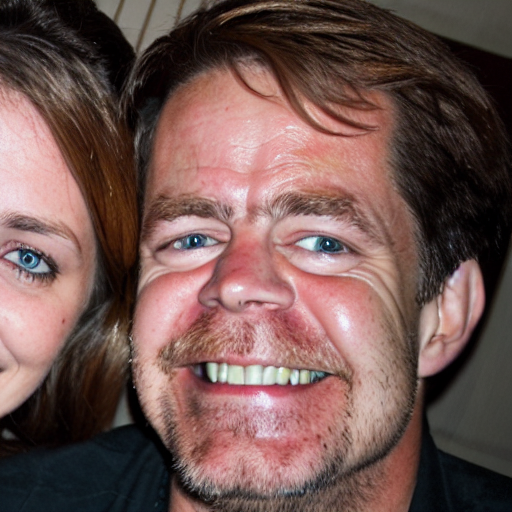}
		\caption*{0.818}
	\end{subfigure}\hfil
	\begin{subfigure}[b]{0.22\linewidth}
		\centering
		\includegraphics[page=1,width=0.95\linewidth, trim={0.1cm 0cm 0cm 0cm},clip]{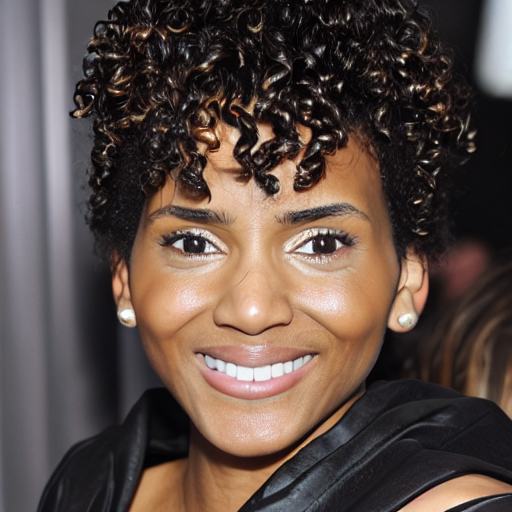}
		\caption*{0.826}
	\end{subfigure}\hfil
	\begin{subfigure}[b]{0.22\linewidth}
		\centering
		\includegraphics[page=1,width=0.95\linewidth, trim={0.1cm 0cm 0cm 0cm},clip]{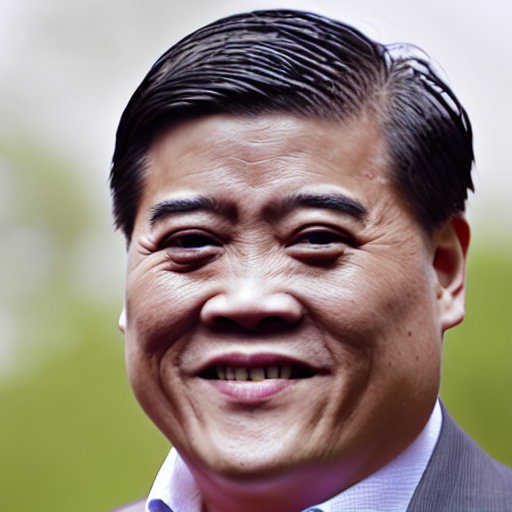}
		\caption*{0.824}
	\end{subfigure}\hfil
	\begin{subfigure}[b]{0.22\linewidth}
		\centering
		\includegraphics[page=1,width=0.95\linewidth, trim={0.1cm 0cm 0cm 0cm},clip]{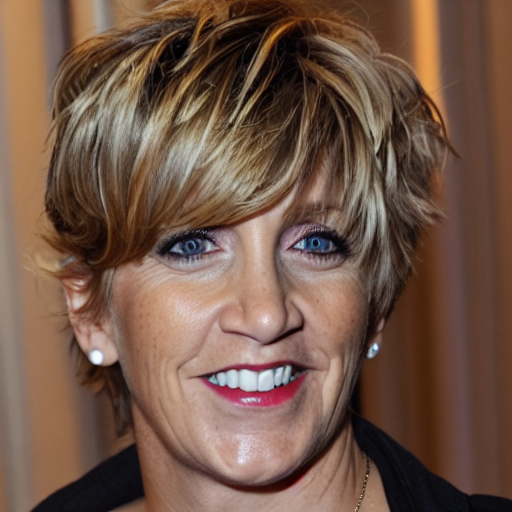}
		\caption*{0.828}
	\end{subfigure}\hfil
	\\ \vspace{3pt}
	\rotatebox[]{90}{\small  Swin \hspace{-70 pt}}\hspace{5pt}\hfil
	\begin{subfigure}[b]{0.22\linewidth}
		\centering
		\includegraphics[page=1,width=0.95\linewidth, trim={0.1cm 0cm 0cm 0cm},clip]{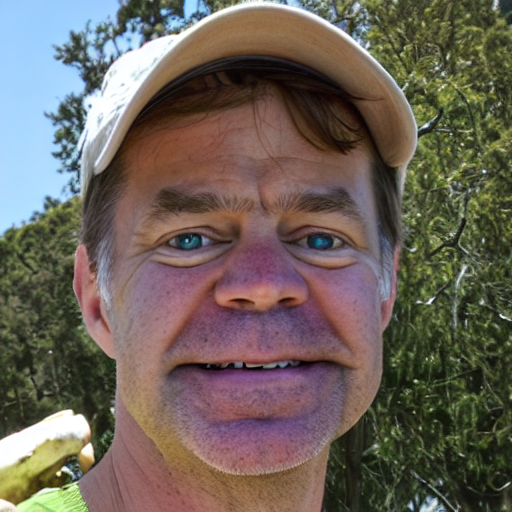}
		\caption*{0.837}
	\end{subfigure}\hfil
	\begin{subfigure}[b]{0.22\linewidth}
		\centering
		\includegraphics[page=1,width=0.95\linewidth, trim={0.1cm 0cm 0cm 0cm},clip]{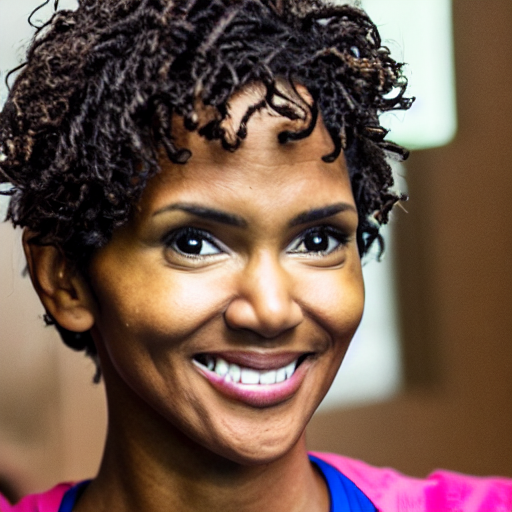}
		\caption*{0.770}
	\end{subfigure}\hfil
	\begin{subfigure}[b]{0.22\linewidth}
		\centering
		\includegraphics[page=1,width=0.95\linewidth, trim={0.1cm 0cm 0cm 0cm},clip]{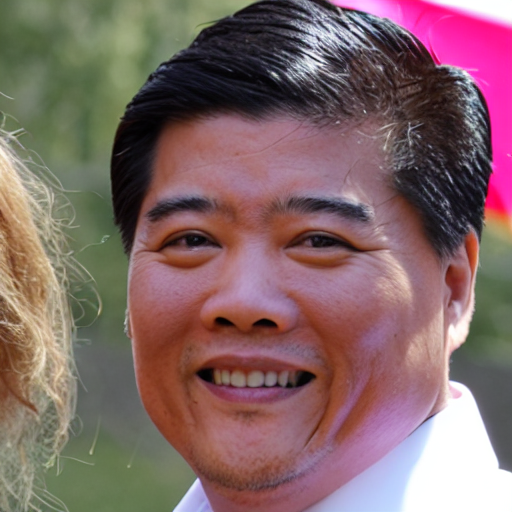}
		\caption*{0.762}
	\end{subfigure}\hfil
	\begin{subfigure}[b]{0.22\linewidth}
		\centering
		\includegraphics[page=1,width=0.95\linewidth, trim={0.1cm 0cm 0cm 0cm},clip]{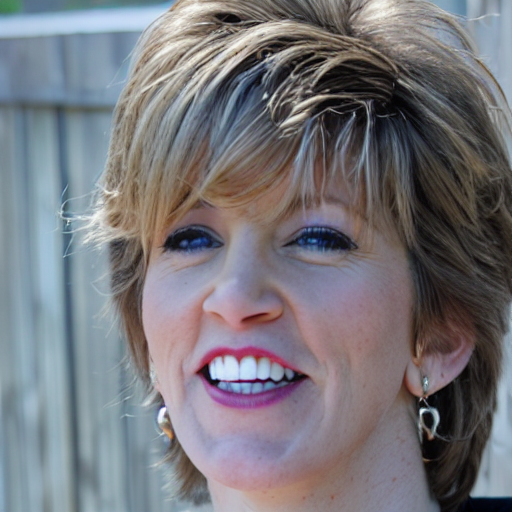}
		\caption*{0.787}
	\end{subfigure}\hfil
	\\ 
	\caption{Sample face images from the dataset (first row) and their corresponding reconstructed face images using various methods. The values show the cosine similarity between embeddings of original and reconstructed face images.}
	\label{fig:exp:sample:success}
\end{figure}

\paragraph{Blackbox Attack}
We train adapter modules for different face recognition models and use the adapter module along with the foundation model to reconstruct face images from embeddings of the MOBIO, LFW, and AgeDB datasets. \cref{tab:exp:blackbox} compares the performance of reconstructed face images with previous methods in the literature in blackbox face reconstruction attacks against different face recognition models.  As the results in this table show our method outperforms previous methods in the literature for different face recognition models and on different datasets. 
Comparing the results with the recognition performance in \cref{tab:exp:FR-performance}, we can observe that a face recognition model with better accuracy is more vulnerable to face reconstruction attacks.
\cref{fig:exp:sample:success} presents sample reconstructed face images in blackbox attacks against different face recognition models using our method. The reconstructed images are of high quality and accurately match the identities of the original images.

\begin{table}[t]
	\renewcommand{\arraystretch}{1.025}
	\setlength{\tabcolsep}{1.5pt}
	\caption{Evaluation of \textit{blackbox} attacks against face recognition models at FMR=$10^{-3}$ on the  MOBIO, LFW, and AgeDB datasets in terms of success attack rate (SAR). In this experiment, the target face recognition model is the same model from which the embeddings are leaked (i.e., $F_\text{victim}=F_\text{target}$).
	}\label{tab:exp:blackbox} 
	\scalebox{0.7725}{
		\begin{tabular}{lcccccccccccccc}
			
   \multirow{2}{*}{\scalebox{0.80}{\textbf{Dataset}}} 
   % \parbox[t]{2mm}{\multirow{2}{*}{\rotatebox[origin=c]{90}{Dataset}}}
   & \multirow{2}{*}{\textbf{Method}} & \multicolumn{6}{c}{\textbf{Victim Face Recognition}}         \\
			\cline{3-8}
			& & \scalebox{0.825}{\textbf{ArcFace}} & \scalebox{0.90}{\textbf{El.Face}} & \scalebox{0.90}{\textbf{Att.Net}}& \scalebox{0.90}{\textbf{HRNet}} & \scalebox{0.80}{\textbf{RepVGG}} & \scalebox{0.90}{\textbf{Swin}}  \\ \toprule
			% \multirow{14}{*}{\textbf{MOBIO}} 
            \parbox[t]{2mm}{\multirow{14}{*}{\rotatebox[origin=c]{90}{MOBIO}}}
            & NBNetA-M~\cite{TPAMI2018reconstruction} &   2.86 & 10.0 & 4.76 & 4.76 & 6.19 & 6.67   \\  \cline{2-8}
			
			&   NBNetA-P~\cite{TPAMI2018reconstruction}  &   23.81 & 60.95 & 15.24 & 14.29 & 44.76 & 30.48 \\  \cline{2-8}
			&   NBNetB-M~\cite{TPAMI2018reconstruction} &    20.95 & 30.0 & 21.43 & 25.24 & 21.43 & 27.62 \\  \cline{2-8}
			&   NBNetB-P~\cite{TPAMI2018reconstruction} &   49.05 & 70.95 & 66.67 & 64.76 & 51.43 & 71.43 \\  \cline{2-8}
			&   BIOSIG 2021~\cite{dong2021towards}	&  24.29 & 34.76 & 38.57 & 16.19 & 24.76 & 18.10 \\  \cline{2-8}
			&   NeurIPSW 2021~\cite{vendrow2021realistic}    &  69.52 & 74.29 & 55.71 & 43.81 & 39.52 & 70.00 \\  \cline{2-8}
			&   \scalebox{0.90}{Com. \& Sec. 2023}~\cite{dong2023reconstruct}	&  87.62 & 90.95 & 80.48 & 71.90 & 44.29 & 82.38 \\  \cline{2-8}

			&   IJCB 2023 \cite{ijcb2023faceti}	&   {80.00}  &  {86.67} &  {81.90} &  {85.24} &  {70.95} &  {84.76}	\\ \cline{2-8}

			&   NeurIPS 2023	\cite{neurips2023faceti} 	&    {81.90} &   {89.05} &   {81.43} &   {86.19} &  {58.10} &   {90.48}  			\\ \cline{2-8}
			
			&   GaFaR	\cite{tpami2023faceti3d}	& 47.62 & 84.76 & 72.38 & 76.67 & 72.86 & 89.05 	\\ \cline{2-8}
			&   GaFaR + GS 	\cite{tpami2023faceti3d}		&  {64.76} &  {86.62} &  {80.00} &  {83.80} &  {73.33} &  {93.33} 		\\ \cline{2-8}
			
			&   ICIP 2023 \cite{shahreza2023blackbox}	&    {95.71} &   {98.10} &   {92.38} &   {97.62} &  85.71 &   {99.52} 			\\ \cline{2-8}
			
			&   \textbf{[Ours]}	 &	\textbf{100.0} & \textbf{99.05} & \textbf{99.52} & \textbf{99.52} & \textbf{98.57} & \textbf{99.52}\\ \midrule

			% \multirow{14}{*}{\textbf{LFW}} 
            \parbox[t]{2mm}{\multirow{14}{*}{\rotatebox[origin=c]{90}{LFW}}}
                &  NBNetA-M~\cite{TPAMI2018reconstruction}  & 14.30 & 37.13 & 10.37 & 20.19 & 10.64 & 13.18  \\  \cline{2-8}
			&   NBNetA-P~\cite{TPAMI2018reconstruction}  & 35.61 & 60.05 & 6.80 & 16.83 & 26.44 & 25.92 \\  \cline{2-8}
			&   NBNetB-M~\cite{TPAMI2018reconstruction} &  26.91 & 52.99 & 17.62 & 31.74 & 18.18 & 27.00 \\  \cline{2-8}
			&   NBNetB-P~\cite{TPAMI2018reconstruction} &  61.66 & 81.74 & 43.42 & 56.30 & 38.12 & 61.02 \\  \cline{2-8}
			&   BIOSIG 2021~\cite{dong2021towards}	& 28.21 & 34.56 & 19.17 & 24.87 & 14.76 & 26.62 \\  \cline{2-8}
			&   NeurIPSW 2021~\cite{vendrow2021realistic}    & 77.00 & 79.37 & 46.52 & 49.52 & 22.4 & 66.07 \\  \cline{2-8}
			&   \scalebox{0.90}{Com. \& Sec. 2023}~\cite{dong2023reconstruct}	&  {87.26} & 89.00 & 55.40 & 59.46 & 28.60 & 69.07 \\  \cline{2-8}

			&   IJCB 2023 \cite{ijcb2023faceti}	&   71.31 &  {81.70} &  {43.58} &  {50.04} &  {35.75} &  {66.57} \\ \cline{2-8}

			&   NeurIPS 2023	\cite{neurips2023faceti} 	&   {77.13}	&   {83.43} &   {45.02} &   {48.22} &  {24.34} &  {61.95} \\ \cline{2-8}
			
			&   GaFaR	\cite{tpami2023faceti3d}	& 51.78	& 74.54 & 33.59 & 37.80 & 25.40 & 67.11  \\ \cline{2-8}
			&   GaFaR + GS 	\cite{tpami2023faceti3d} &  {61.56} &  {78.67} &  {38.42} &  {43.27} &  {29.84} &  {70.82} \\ \cline{2-8}
			
			&   ICIP 2023 \cite{shahreza2023blackbox}	&    {90.67}	&   {92.27} &   {62.20} &   {68.99} & 44.56 &  {82.21}  \\ \cline{2-8}
			
			&   \textbf{[Ours]}	 &	\textbf{95.71} & \textbf{93.18} & \textbf{78.71} & \textbf{80.66} & \textbf{67.11} & \textbf{88.44}\\ 
            \midrule

\parbox[t]{2mm}{\multirow{14}{*}{\rotatebox[origin=c]{90}{AgeDB}}} &
             			NBNetA-M~\cite{TPAMI2018reconstruction}  &    0.81 & 2.55 & 0.22 & 0.38 & 0.44 & 0.27 \\ \cline{2-8}
 		&	NBNetA-P~\cite{TPAMI2018reconstruction}   &     3.99 & 8.92 & 0.34 & 0.14 & 3.71 & 1.02\\ \cline{2-8}
 		&	NBNetB-M~\cite{TPAMI2018reconstruction}  &      1.88 & 6.27 & 0.50 & 0.77 & 1.06 & 0.68   \\ \cline{2-8}
 		&	NBNetB-P~\cite{TPAMI2018reconstruction}  &     13.18 & 28.94 & 5.08 & 5.61 & 7.92 & 8.75 \\ \cline{2-8}
 		&	BIOSIG 2021~\cite{dong2021towards}	&  3.93 & 4.88 & 1.58 & 1.97 & 2.22 & 2.48  \\ \cline{2-8}
 		&	NeurIPSW 2021~\cite{vendrow2021realistic}   & {29.64} & 34.89 & 15.06 & 12.02   & 14.49 & 21.10   \\ \cline{2-8}
            &	\scalebox{0.90}{Com. \& Sec. 2023}~\cite{dong2023reconstruct}  & {58.80}  & 66.10  & 36.82  & 32.45  & 14.98  & 37.81 \\ \cline{2-8}
            &   IJCB 2023 \cite{ijcb2023faceti}	&  {57.94}  & {76.98}  & {49.92}  & {50.57}  & {48.36}  & {59.95} \\ \cline{2-8}
            & NeurIPS 2023	\cite{neurips2023faceti} & 48.57  & 64.89  & 26.26  & 29.39  & 19.15  & 30.63 \\ \cline{2-8}
 		&	GaFaR	\cite{tpami2023faceti3d}		&  21.67	& 47.37 & 14.59 & 17.09 & 18.02 & 30.05	\\\cline{2-8}
 		&	GaFaR + GS	\cite{tpami2023faceti3d}	&  28.94 & 53.10 & 18.76 & 22.40 & 24.01 & 35.20    	\\\cline{2-8}
        &   ICIP 2023 \cite{shahreza2023blackbox}	& 58.07   & 79.20 & 46.21 & 46.75 & 52.29 & 62.75 \\ \cline{2-8}
        &   \textbf{[Ours]}	 &	\textbf{86.42} & \textbf{85.64} & \textbf{84.31} & \textbf{78.43} & \textbf{79.43} & \textbf{85.69} \\
   \bottomrule
		\end{tabular}
	}

\end{table}

\begin{table}[t]
	\renewcommand{\arraystretch}{1.025}
	\setlength{\tabcolsep}{2.85pt}
	\caption{Transferability evaluation (i.e., $F_\text{victim}\neq F_\text{target}$) of reconstructed face images from \textit{blackbox} attacks against ArcFace embeddings ($F_\text{victim}$) for entering a \textit{different target}  face recognition system ($F_\text{target}$) at FMR=$10^{-3}$ on the  MOBIO, LFW, and AgeDB datasets in terms of success attack rate (SAR). 
	}\label{tab:exp:transferability} 
	\scalebox{0.7725}{
		\begin{tabular}{lcccccccccc}
			
   \multirow{2}{*}{\scalebox{0.80}{\textbf{Dataset}}} 
   % \parbox[t]{2mm}{\multirow{2}{*}{\rotatebox[origin=c]{90}{Dataset}}}
   & \multirow{2}{*}{\textbf{Method}} & \multicolumn{5}{c}{\textbf{Target Face Recognition}}         \\
			\cline{3-7}
			& & \scalebox{0.90}{\textbf{El.Face}} & \scalebox{0.90}{\textbf{Att.Net}}& \scalebox{0.90}{\textbf{HRNet}} & \scalebox{0.80}{\textbf{RepVGG}} & \scalebox{0.90}{\textbf{Swin}}  \\ \toprule
			% \multirow{14}{*}{\textbf{MOBIO}} 
            \parbox[t]{2mm}{\multirow{14}{*}{\rotatebox[origin=c]{90}{MOBIO}}}
            & NBNetA-M~\cite{TPAMI2018reconstruction} &  0.0   &  0.0   &  0.0   &  0.0   &  0.0        \\  \cline{2-7}
			
			&   NBNetA-P~\cite{TPAMI2018reconstruction}  &  1.43 & 2.86 & 1.43 & 1.90 & 0.95   \\ \cline{2-7}
			&   NBNetB-M~\cite{TPAMI2018reconstruction} &   4.29 & 3.81 & 4.29 & 1.90 & 4.29    \\ \cline{2-7}
			&   NBNetB-P~\cite{TPAMI2018reconstruction} &   10.95 & 6.67 & 6.19 & 2.38 & 13.33  \\ \cline{2-7}
			&   BIOSIG 2021~\cite{dong2021towards}	&  3.81 & 2.86 & 3.33 & 2.38 & 4.29  \\ \cline{2-7}
			&   NeurIPSW 2021~\cite{vendrow2021realistic}    & 17.14 & 5.71 & 10.48 & 3.81 & 13.81  \\ \cline{2-7}
			&   \scalebox{0.90}{Com. \& Sec. 2023}~\cite{dong2023reconstruct}	&  37.61 & 26.67 &  30.48 &  19.52  & 40.00 \\ \cline{2-7}

			&   IJCB 2023 \cite{ijcb2023faceti}	&  74.76 & 59.52 & 59.52 & 55.24 & 68.57  \\ \cline{2-7}

			&   NeurIPS 2023	\cite{neurips2023faceti} 	&   73.81 & 54.29 & 57.14 &  44.29  & 67.14   \\ \cline{2-7}
			
			&   GaFaR	\cite{tpami2023faceti3d}	&  52.38 & 29.05 & 31.90 & 28.10 & 43.33 \\ \cline{2-7}
			&   GaFaR + GS 	\cite{tpami2023faceti3d}		&  55.24 & 36.19 & 41.90 & 32.86 & 50.00  \\ \cline{2-7}
			
			&   ICIP 2023 \cite{shahreza2023blackbox}	&  90.48 & 73.33 & 75.24 & 61.43 & 72.38   \\ \cline{2-7}
			
			&   \textbf{[Ours]}	 &  \textbf{99.52}	&  \textbf{97.62} & \textbf{97.62} & \textbf{84.29}  &  \textbf{98.57}    \\ \midrule

			% \multirow{14}{*}{\textbf{LFW}} 
            \parbox[t]{2mm}{\multirow{14}{*}{\rotatebox[origin=c]{90}{LFW}}}
                &  NBNetA-M~\cite{TPAMI2018reconstruction}  &  3.32 & 0.51 & 0.41 & 0.23 & 1.60    \\  \cline{2-7}
			
			&   NBNetA-P~\cite{TPAMI2018reconstruction}  &  8.56 & 1.22 & 1.44 & 0.76 & 4.79   \\ \cline{2-7}
			&   NBNetB-M~\cite{TPAMI2018reconstruction} &  13.09 & 1.89 & 2.03 & 0.86 & 8.44    \\ \cline{2-7}
			&   NBNetB-P~\cite{TPAMI2018reconstruction} &   34.41 & 7.21 & 7.77 & 4.06 & 23.82  \\ \cline{2-7}
			&   BIOSIG 2021~\cite{dong2021towards}	&    6.25 & 2.12 &  1.70 & 1.39 & 6.22 \\ \cline{2-7}
			&   NeurIPSW 2021~\cite{vendrow2021realistic}    &  29.06 & 9.79 & 9.51 & 4.48 & 20.75  \\ \cline{2-7}
			&   \scalebox{0.90}{Com. \& Sec. 2023}~\cite{dong2023reconstruct}	& 50.47 & 24.16 & 24.72 &  13.99  & 41.27  \\ \cline{2-7}

			&   IJCB 2023 \cite{ijcb2023faceti}	&  62.33 & 25.59 &  25.91 & 14.72 & 44.91   \\ \cline{2-7}

			&   NeurIPS 2023	\cite{neurips2023faceti} 	&   68.05 & 28.87 & 28.45 &  16.90  & 28.28   \\ \cline{2-7}
			
			&   GaFaR	\cite{tpami2023faceti3d}	&  43.59 & 14.58 & 14.77 & 8.42 & 29.69 \\ \cline{2-7}
			&   GaFaR + GS 	\cite{tpami2023faceti3d}		&   47.57 & 17.13 & 17.35 & 10.11 & 33.16 \\ \cline{2-7}
			
			&   ICIP 2023 \cite{shahreza2023blackbox}	&  51.93 & 87.28  & 56.62 & 31.23 & 72.73 \\ \cline{2-7}
			
			&   \textbf{[Ours]}	 &  \textbf{92.40}	 &  \textbf{73.40}	 &  \textbf{76.24}	 &  \textbf{52.54}	 &  \textbf{86.70}  \\ \midrule

\parbox[t]{2mm}{\multirow{14}{*}{\rotatebox[origin=c]{90}{AgeDB}}} &
             			NBNetA-M~\cite{TPAMI2018reconstruction}  &     0.73 & 0.04 & 0.12 & 0.20 & 0.10   \\  \cline{2-7}
			
			&   NBNetA-P~\cite{TPAMI2018reconstruction}  &  2.84 & 0.57 & 0.41 & 0.87 & 0.72    \\ \cline{2-7}
			&   NBNetB-M~\cite{TPAMI2018reconstruction} &   2.21 & 0.53 & 0.67 & 0.70 & 0.52  \\ \cline{2-7}
			&   NBNetB-P~\cite{TPAMI2018reconstruction} &   13.15 & 3.92 & 3.72 & 4.13 & 5.32  \\ \cline{2-7}
			&   BIOSIG 2021~\cite{dong2021towards}	& 2.30 & 1.07 & 1.08 & 1.24 & 1.33   \\ \cline{2-7}
			&   NeurIPSW 2021~\cite{vendrow2021realistic}    & 17.74 & 8.19 & 7.84 & 8.14 & 9.80  \\ \cline{2-7}
			&   \scalebox{0.90}{Com. \& Sec. 2023}~\cite{dong2023reconstruct}	&  30.52 & 17.26 & 16.27 & 18.79 & 18.87 \\ \cline{2-7}

			&   IJCB 2023 \cite{ijcb2023faceti}	&  39.39 & 16.11 & 15.29 & 18.09 & 21.83   \\ \cline{2-7}

			&   NeurIPS 2023	\cite{neurips2023faceti} 	&  41.30 & 16.10 & 14.77 & 17.33 & 22.04    \\ \cline{2-7}
			
			&   GaFaR	\cite{tpami2023faceti3d}	& 20.13 & 6.24 & 5.94 & 6.60 & 8.08  \\ \cline{2-7}
			&   GaFaR + GS 	\cite{tpami2023faceti3d} & 23.85 & 8.17 & 8.07 & 9.51 & 10.57   \\ \cline{2-7}
			
			&   ICIP 2023 \cite{shahreza2023blackbox}	& 58.44 & 37.78  & 32.28 &  42.15  & 41.56 \\ \cline{2-7}
			
			&   \textbf{[Ours]}	 &	\textbf{80.92} & \textbf{74.48} & \textbf{67.77} & \textbf{75.16} & \textbf{78.33}    \\
   \bottomrule
		\end{tabular}
	}

\end{table}

\paragraph{Transferability Evaluation}
To evaluate the transferability of reconstructed face images, we use the generated images to enter a different face recognition system (i.e., $F_\text{victim}\neq F_\text{target}$). \cref{tab:exp:transferability} reports the transferability using reconstructed face images from ArcFace embeddings (i.e., $F_\text{victim}$) when attacking a \textit{different} target face recognition system ($F_\text{target}$), and compares it with different methods in the literature. As the results in this table show, the reconstructed face images are transferable and can be also used to enter a different face recognition system. The results in this table also show that our method outperforms previous methods in transferability evaluation and demonstrates the effectiveness of our reconstruction attack. Comparing the results in \cref{tab:exp:transferability} with our evaluation of blackbox attacks in \cref{tab:exp:blackbox}, we observe that for all methods, the transferability evaluation leads to a lower success rate. 
We would like to stress that the transferability of reconstructed images simulates a very difficult attack scenario and further sheds light on the vulnerability of face recognition systems.

\begin{table}[t]
	\renewcommand{\arraystretch}{1.025}
	\setlength{\tabcolsep}{3.5pt}
	\caption{Ablation study on the performance of adapter module when trained with different numbers of training images and used in \textit{blackbox} attacks against different face recognition models at FMR=$10^{-3}$ on the  LFW  dataset in terms of success attack rate (SAR). In this experiment, the target model is the same model from which the embeddings are leaked. The training time of the adapter module is also reported in seconds.
	}\label{tab:exp:num-images} 
	\scalebox{0.75}{
		\begin{tabular}{ccccccccccccccc}
			
   \textbf{Number }
   & \textbf{Training} & \multicolumn{6}{c}{\textbf{Face Recognition}}         \\
			\cline{3-8}
		\textbf{of Images}	& \textbf{Time (sec)} & \scalebox{0.85}{\textbf{ArcFace}} & \scalebox{0.90}{\textbf{El.Face}} & \scalebox{0.90}{\textbf{Att.Net}}& \scalebox{0.90}{\textbf{HRNet}} & \scalebox{0.80}{\textbf{RepVGG}} & \scalebox{0.90}{\textbf{Swin}}  \\ \toprule
500 & 0.32    & 89.30 & 80.2 & 57.40 & 48.04 & 43.70 & 73.79 \\ \hline
600 & 0.38    & 90.78 & 82.88 & 64.01 & 56.88 & 50.07 & 78.59 \\ \hline
800 & 0.49    & 92.66 & 86.28 & 70.64 & 64.92 & 55.36 & 82.35 \\ \hline
1000 & 0.63     & 93.60 & 88.27 & 73.12 & 69.59 & 57.80 & 84.25 \\ \hline
2000 & 1.23    & 94.82 & 91.41 & 77.02 & 77.52 & 63.03 & 87.25 \\ \hline
5000 & 3.05    & 95.40 & 92.18 & 78.28 & 80.41 & 65.45 & 88.32 \\ \hline
10000 & 6.05    & 95.69 & 92.95 & 78.53 & 80.55 & 66.71 & 87.97 \\ \hline
15000 & 9.10    & 95.69 & 92.49 & 78.32 & 80.64 & 66.86 & 88.45 \\ %\hline
   \bottomrule
		\end{tabular}
	}

\end{table}

\paragraph{Training Adapter Module and the Effect of Number of Images 
}
Compared to other methods which require long training (learning-based methods) or inference (optimization-based methods) time, our method relies on training a light adapter module to use a foundation model for face reconstruction attacks.
Our approach for training the adapter module involves estimating a linear layer with dimensions\footnote{Note that face recognition models used in our experiment as well as the one for foundation model $F_\text{FM}$ have embeddings with 512 dimensions. However, a face recognition model with a different embedding dimension can still be used and requires training an adapter module with corresponding dimensions.} of $512\times512$. Once we obtain the embeddings from the $F_\text{FM}$ model and the victim model from a number of images, the computation required to estimate these parameters is minimal. To assess the model’s performance with varying amounts of training data, we conduct an ablation study by changing the number of training images. For this evaluation, we use images from the FFHQ dataset 
and evaluate the performance of adapters trained with different quantities of training samples in face reconstruction attacks. 
%We select 10,000 samples as the test set and evaluate the performance of models trained with different quantities of training samples. 
% These models are then evaluated on the LFW dataset, and the SARs are reported in Table XX.
\cref{tab:exp:num-images} compares the performance of adapter modules trained with different numbers of images when used in blackbox attacks against different face recognition models. 
As the results in this table show with more number of images the performance of the adapter module increases. However, 10,000 images are almost enough to train the adapter module with close to maximum performance.  Compared with previous methods in \cref{tab:exp:blackbox} with only 600 training samples, the adapter module can outperform previous methods in the literature.
This table also reports the required time to train the adapter module for different numbers of images on a system equipped with an NVIDIA GeForce RTX\textsuperscript{TM} 3090. In fact, in contrast to the optimization methods, the training of the adapter module is required to be conducted only once for each model, and then the adversary can use the trained adapter to reconstruct unlimited leaked embeddings in a fraction of a second. Compared to previous learning-based methods, the training time for our adapter module is almost negligible to the training time of all previous learning-based methods.

\begin{figure}[tbh] 
	\centering
	\rotatebox[]{90}{\small  Original \hspace{-45 pt}}\hspace{5pt}\hfil
	\begin{subfigure}[b]{0.22\linewidth}
		\centering
		\includegraphics[page=1,width=.95\linewidth, trim={0.1cm 0cm 0cm 0cm},clip]{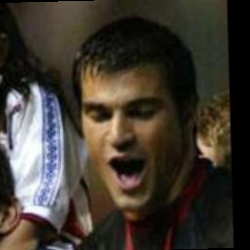}
	\end{subfigure}\hfil
	\begin{subfigure}[b]{0.22\linewidth}
		\centering
		\includegraphics[page=1,width=.95\linewidth, trim={0.1cm 0cm 0cm 0cm},clip]{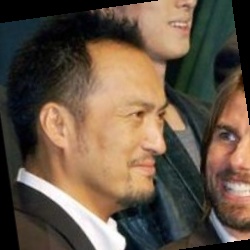}
	\end{subfigure}\hfil
	\begin{subfigure}[b]{0.22\linewidth}
		\centering
		\includegraphics[page=1,width=.95\linewidth, trim={0.1cm 0cm 0cm 0cm},clip]{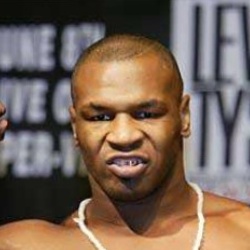}
	\end{subfigure}\hfil
	\begin{subfigure}[b]{0.22\linewidth}
		\centering
		\includegraphics[page=1,width=.95\linewidth, trim={0.1cm 0cm 0cm 0cm},clip]{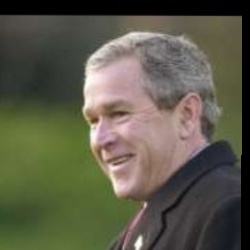}
	\end{subfigure}\hfil
	\\ \vspace{3pt}
	\rotatebox[]{90}{\small  ArcFace \hspace{-70 pt}}\hspace{5pt}\hfil
	\begin{subfigure}[b]{0.22\linewidth}
		\centering
		\includegraphics[page=1,width=0.95\linewidth, trim={0.1cm 0cm 0cm 0cm},clip]{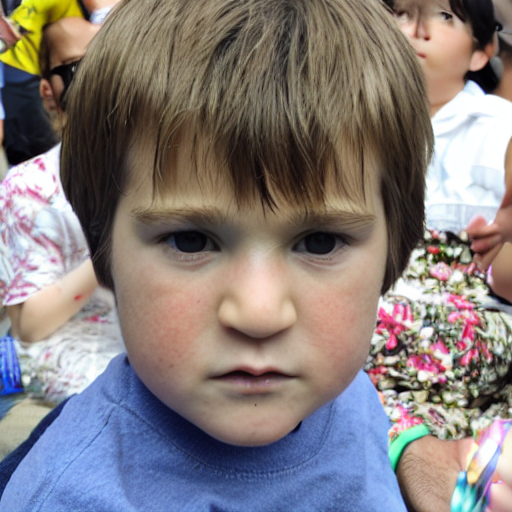}
		\caption*{0.021}
	\end{subfigure}\hfil
	\begin{subfigure}[b]{0.22\linewidth}
		\centering
		\includegraphics[page=1,width=0.95\linewidth, trim={0.1cm 0cm 0cm 0cm},clip]{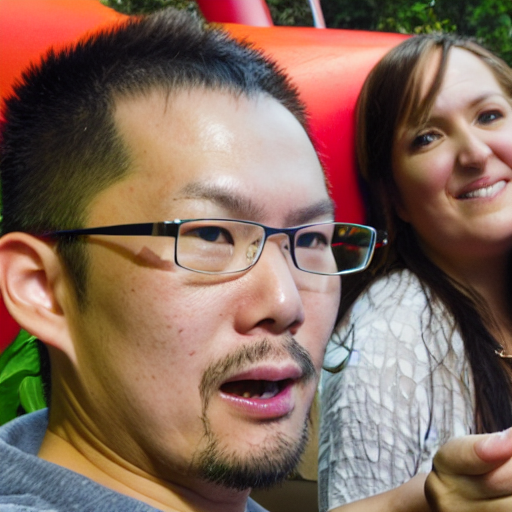}
		\caption*{0.118}
	\end{subfigure}\hfil
	\begin{subfigure}[b]{0.22\linewidth}
		\centering
		\includegraphics[page=1,width=0.95\linewidth, trim={0.1cm 0cm 0cm 0cm},clip]{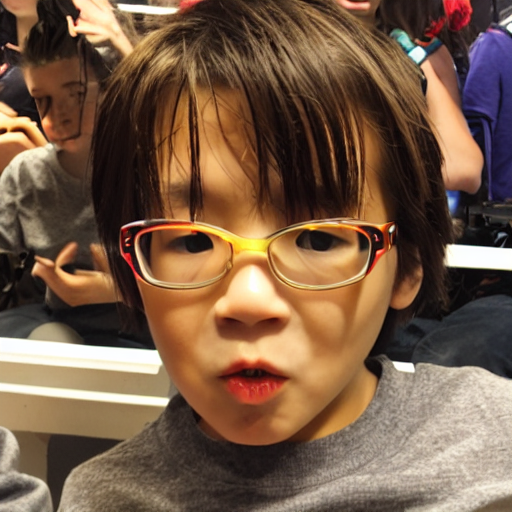}
		\caption*{0.310}
	\end{subfigure}\hfil
	\begin{subfigure}[b]{0.22\linewidth}
		\centering
		\includegraphics[page=1,width=0.95\linewidth, trim={0.1cm 0cm 0cm 0cm},clip]{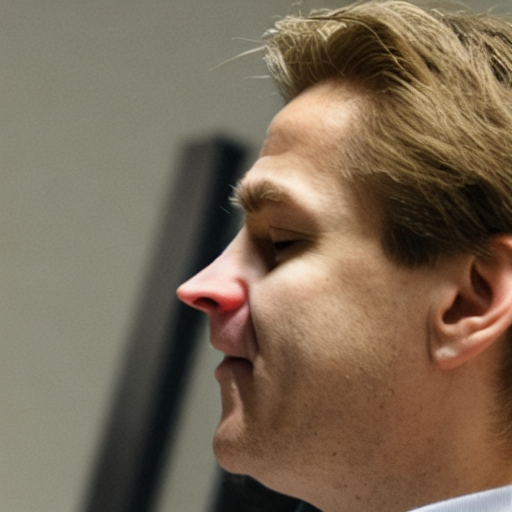}
		\caption*{0.357}
	\end{subfigure}\hfil
	\\ 
	\caption{Sample failure cases:  Images from the LFW dataset (first row) and their corresponding reconstructed face images in blackbox attack against ArcFace. The values show the cosine similarity between embeddings of original and reconstructed face images.}
	\label{fig:exp:sample:failure}
\end{figure}

\begin{figure}[tbh]
	\centering

	% \rotatebox[]{90}{\small  RepVGG\_B1 \hspace{-70 pt}}\hspace{5pt}\hfil
	\begin{subfigure}[b]{0.22\linewidth}
		\centering
		\includegraphics[page=1,width=0.95\linewidth, trim={0.1cm 0cm 0cm 0cm},clip]{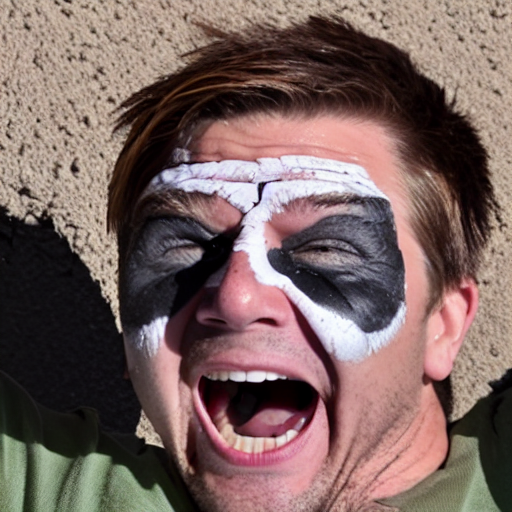}
		% \caption*{0.297}
	\end{subfigure}\hfil
	\begin{subfigure}[b]{0.22\linewidth}
		\centering
		\includegraphics[page=1,width=0.95\linewidth, trim={0.1cm 0cm 0cm 0cm},clip]{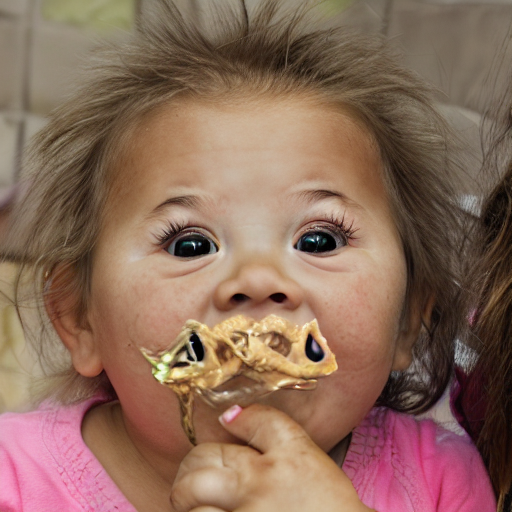}
		% \caption*{0.252}
	\end{subfigure}\hfil
	\begin{subfigure}[b]{0.22\linewidth}
		\centering
		\includegraphics[page=1,width=0.95\linewidth, trim={0.1cm 0cm 0cm 0cm},clip]{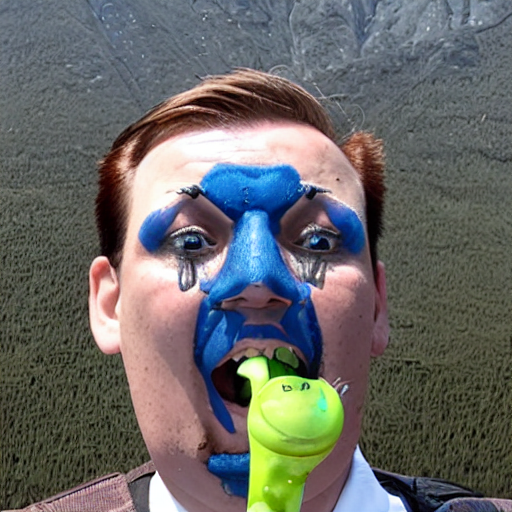}
		% \caption*{0.413}
	\end{subfigure}\hfil
	\begin{subfigure}[b]{0.22\linewidth}
		\centering
		\includegraphics[page=1,width=0.95\linewidth, trim={0.1cm 0cm 0cm 0cm},clip]{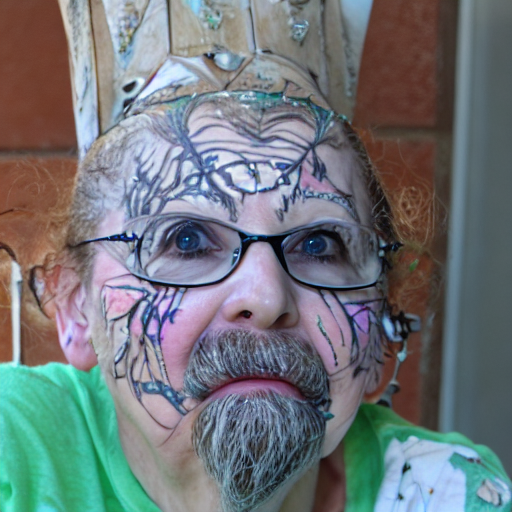}
		% \caption*{0.264}
	\end{subfigure}\hfil
	\\ 
	\caption{Samples of reconstructed face images with artifacts.}
	\label{fig:exp:sample:failures-odd}
\end{figure}

\section{Discussion}\label{sec:discussion}

\paragraph{Failure Cases}
\cref{fig:exp:sample:success} presents samples that have been successfully reconstructed and identified with a high match score as the original identities. Conversely, \cref{fig:exp:sample:failure} illustrates some failure cases of our approach in blackbox attack against ArcFace embeddings. 
Although the model attempts to preserve the identity, it does not constrain the age, leading to inconsistencies. The model also struggles with extreme expressions and poses, and in some instances, it fails to preserve the ethnicity of the original samples. While the reconstructions are effective in terms of attack success rates, these limitations suggest that the face embedding may not capture sufficient details for certain attributes.

Another group of failure cases corresponds to images that have artifacts in the generation using the Arc2Face model. \cref{fig:exp:sample:failures-odd} illustrates such cases, where the generated face images do not look realistic. The generation of these samples can be caused by an error in translating embeddings by our adapter module. In such a case, the generated embedding by our adapter module may not fit the distribution of embedding space of the foundation model, and therefore the generator may generate out of distribution image.

\paragraph{Societal Impacts} 
In this paper, we proposed an efficient adapter module to map the embeddings of any face recognition model to the embeddings space of Arc2Face as a foundation model. The reconstructed face images by our method have high quality and outperform the state-of-the-art in the literature, demonstrating the effectiveness of our adapter module. Our adapter module can be used in future foundation models which are based on embeddings of a preset face recognition model. 
Our experiments also show that the reconstructed face images by our method using a foundation can achieve high success attack rates when attacking a face recognition system. The reconstructed face images are also transferable and can be used to enter any other system in which the subjects are enrolled. Our results also shed light on the critical vulnerability of face recognition systems to face reconstruction attacks. While this work is conducted with the motivation of showing the vulnerability of face recognition systems, we do not condone using our method with the intent of attacking a real face recognition system. 
In a similar vein, data protection regulations, e.g.,  the European Union General Data Protection Regulation (EU-GDPR)~\cite{GDPR}, classify biometric data as sensitive information, and impose legal obligations to protect them. To address and mitigate such attacks against biometric systems, several template protection methods are proposed in the literature~\cite{nandakumar2015biometric,kaur2022biometric,kumar2020cancelable,otroshi2024information,shahreza2022hybrid,shahreza2023mlp}. 
We should also note that the project on which the work has been conducted has undergone and passed an {Internal Ethical Review Board (IRB)}.

\section{Conclusion}\label{sec:conclusion}
The foundation models are trained with a large volume of data and can be used in different tasks.
In this paper, we used a face foundation model and proposed a new face reconstruction attack against face recognition systems. While the foundation model can generate face images given embeddings of a preset face recognition model, we trained an efficient adapter module that can map embeddings of a blackbox face recognition model to the embedding space of the foundation model. Then, we can use the foundation model to generate the face images from embeddings of various face recognition models. Our experiments show that our method outperforms previous face reconstruction methods in the literature and demonstrate the effectiveness of our approach. In particular, the adversary requires a limited computation resource to train the adapter module and then conduct the face reconstruction attack. The reconstructed face images by our method are realistic and have high quality. In addition, the reconstructed images are transferable and can be used to enter any other system in which the same subject is enrolled. 
Our results show the serious vulnerability of face recognition systems to face reconstruction attacks. Moreover, the results achieved by training a light adapter module in conjunction with a pretrained foundation model shed light on the potential of foundation models and demonstrate that the application of a pretrained foundation model can be easily used with different face recognition models. Hence, this paper paves the way for future studies on the application of foundation models for various problems related to face recognition systems.  
%-------------------------------------------------------------------------

\section*{Acknowledgment}
This research is based upon work funded by the Hasler foundation through the Responsible Face Recognition (SAFER) project as well as the H2020 TReSPAsS-ETN Marie Sk\l{}odowska-Curie-Curie early training network under agreement 860813. This work was also supported by the Swiss Center for Biometrics Research \& Testing at Idiap Research Institute.

%%%%%%%%% REFERENCES
{\small
\bibliographystyle{ieee_fullname}
\bibliography{egbib}
}

\end{document}